\documentclass{article}

\usepackage{microtype}
\usepackage{graphicx}
\usepackage{subfigure}
\usepackage{booktabs} 

\usepackage{hyperref}
\usepackage{url}
\usepackage{amsmath, amssymb, amsfonts}
\usepackage{algorithm}
\usepackage{algpseudocode}
\usepackage{array}
\usepackage{bbm}

\usepackage{amsmath,amsfonts,bm}

\def\eqref#1{equation~\ref{#1}}

\def\1{\bm{1}}

\DeclareMathAlphabet{\mathsfit}{\encodingdefault}{\sfdefault}{m}{sl}
\SetMathAlphabet{\mathsfit}{bold}{\encodingdefault}{\sfdefault}{bx}{n}

\usepackage[accepted]{icml2025}
\makeatletter
\renewcommand{\ICML@appearing}{}
\makeatother
\usepackage{mathtools}
\usepackage{amsthm}

\usepackage[capitalize,noabbrev]{cleveref}

\theoremstyle{plain}
\newtheorem{theorem}{Theorem}[section]

\theoremstyle{definition}

\theoremstyle{remark}
\newtheorem{remark}[theorem]{Remark}

\newcommand{\Ours}{GoFlow }
\newcommand{\OursNoSpace}{GoFlow}

\usepackage[textsize=tiny]{todonotes}

\icmltitlerunning{Flow-based Domain Randomization for
Learning and Sequencing Robotic Skills}

\begin{document}

\twocolumn[
\icmltitle{Flow-based Domain Randomization for
Learning and \\ Sequencing Robotic Skills}




\begin{icmlauthorlist}
\icmlauthor{Aidan Curtis}{mit}
\icmlauthor{Eric Li}{mit}
\icmlauthor{Michael Noseworthy}{mit}
\icmlauthor{Nishad Gothoskar}{mit}
\icmlauthor{Sachin Chitta}{autodesk}
\icmlauthor{Hui Li}{autodesk}
\icmlauthor{Leslie Pack Kaelbling}{mit}
\icmlauthor{Nicole Carey}{autodesk}

\end{icmlauthorlist}

\icmlaffiliation{mit}{MIT CSAIL}
\icmlaffiliation{autodesk}{Autodesk Research}

\icmlcorrespondingauthor{Aidan Curtis}{curtisa@mit.edu}

\icmlkeywords{Reinforcement Learning, Planning, Robotics}

\vskip 0.3in
]

\printAffiliationsAndNotice{Work done during a summer internship at Autodesk.} 

\begin{abstract}
Domain randomization in reinforcement learning is an established technique for increasing the robustness of control policies trained in simulation. 
By randomizing environment properties during training, the learned policy can become robust to uncertainties along the randomized dimensions. 
While the environment distribution is typically specified by hand, in this paper we investigate automatically discovering a sampling distribution via entropy-regularized reward maximization of a normalizing-flow–based neural sampling distribution. 
We show that this architecture is more flexible and provides greater robustness than existing approaches that learn simpler, parameterized sampling distributions, as demonstrated in six simulated and one real-world robotics domain. 
Lastly, we explore how these learned sampling distributions, along with a privileged value function, can be used for out-of-distribution detection in an uncertainty-aware multi-step manipulation planner.
\end{abstract}

\section{Introduction}

Reinforcement learning (RL) is a powerful tool in robotics because it can be used to learn effective control policies for systems with highly complex dynamics that are difficult to model analytically. Unlike traditional control methods, which rely on precise mathematical models, RL learns directly from simulated or real-world experience~\cite{luo2021learning, zhu2020ingredients, schoettler2020deep}. 

However, RL approaches can be inefficient, involving slow, minimally parallelized, and potentially unsafe data-gathering processes when performed in real environments~\cite{kober2013reinforcement}. 
Learning in simulation eliminates some of these problems, but introduces new issues in the form of discrepancies between the training and real-world environments~\cite{valassakis2020crossing}.

Successful RL from simulation hence requires efficient and accurate models of both robot and environment during the training process. 
But even with highly accurate geometric and dynamic simulators, the system can still be only considered partially observable ~\cite{kober2013reinforcement}---material qualities, inertial properties, perception noise, contact and force sensor noise, manufacturing deviations and tolerances, and imprecision in robot calibration all add uncertainty to the model. 

To improve the robustness of learned policies against sim-to-real discrepancies, it is common to employ domain randomization, varying the large set of environmental parameters inherent to a task according to a given underlying distribution \cite{muratore2019assessing}. 
In this way, policies are trained to maximize their overall performance over a diverse set of models. 
These sampling distributions are typically constructed manually with Gaussian or uniform distributions on individual parameters with hand-selected variances and bounds.
However, choosing appropriate distributions for each of the domain randomization parameters remains a delicate process ~\cite{josifovski2022analysis}; too broad a distribution leads to suboptimal local minima convergence (see Figure~\ref{fig:sim_results}), while too narrow a distribution leads to poor real-world generalization~\citep{gaussian_dr, DBLP:journals/corr/abs-1810-12282}.
Many existing methods rely on real-world rollouts from hardware experiments to estimate dynamics parameters~\cite{chebotar2019closing, bayessim, pmlr-v164-muratore22a}. However, for complex tasks with physical parameters that are difficult to efficiently or effectively sample, this data may be time-consuming to produce, or simply unavailable.

An alternative strategy is to learn an environment distribution during training with the aim of finding the broadest possible training distribution that can feasibly be solved in order to maximize the chances of transferring to an unknown target environment. Automating updates to parameter distributions during the training process can remove the need for heuristic tuning and iterative experimentation~\cite{gaussian_dr, adr, entmax}. 
In this paper, we present \OursNoSpace, a novel approach for learned domain randomization that combines actor-critic reinforcement learning architectures~\citep{schulman2017proximal, haarnoja2018soft} with a neural sampling distribution to learn robust policies that generalize to real-world settings. 
By maximizing the diversity of parameters during sampling, we actively discover environments that are challenging for the current policy but still solvable given enough training. 

As proof of concept, we investigate one real-world use case: contact-rich manipulation for assembly. 
Assembly is a critical area of research for robotics, requiring a diverse set of high-contact interactions which often involve wide force bandwidths and unpredictable dynamic changes. 
Recently, sim-to-real RL has emerged as a potentially useful strategy for learning robust contact-rich policies without laborious real-world interactions~\cite{forge, industreal, dynamic_compliance}. We build on this work by testing our method on the real-world industrial assembly task of gear insertion.

Lastly, we extend this classical gear insertion task to the setting of multi-step decision making under uncertainty and partial observability. As shown in this paper and elsewhere~\cite{entmax, gaussian_dr, adr}, policies trained in simulation have an upper bound on the environmental uncertainties that they can be conformant to. For example, a visionless robot executing an insertion policy can only tolerate so much in-hand pose error.
However, estimates of this uncertainty can be used to inform high level control decisions through task-oriented information gathering~\cite{task_oriented_exploration, curtis2022taskdirectedexplorationcontinuouspomdps}. For example, looking closer at objects can result in more accurate pose estimates or tracking objects in the hand to detect slippage.
By integrating a probabilistic pose estimation model, we can use the sampling distributions learned with \Ours as an out-of-distribution detector to determine whether the policy is expected to succeed under its current belief about the world state. 
If the robot has insufficient information, it can act to deliberately seek the needed information using a simple belief-space planning algorithm.

Our contributions are as follows: We introduce \OursNoSpace, a novel domain randomization method that combines actor-critic reinforcement learning with a learned neural sampling distribution. We show that \Ours achieves higher domain coverage than fixed and other learning-based solutions to domain randomization on a suite of simulated environments. We demonstrate the efficacy of \Ours in a real-world contact-rich manipulation task—gear insertion—and extend it to multi-step decision-making under uncertainty. By integrating a probabilistic pose estimation model, we enable the robot to actively gather additional information when needed, enhancing performance in partially observable settings.

\section{Related Work}
Recent developments in reinforcement learning have demonstrated that policies trained in simulation can effectively transfer to real-world robots, even for contact-rich robotic tasks~\citep{dynamic_compliance, industreal, forge, Jin2023}. A key innovation enabling this transferability is domain randomization~\citep{understanding_dr, original_dr}, where environment parameters are sampled from predefined distributions during training, enabling learned policies to generalize to environmental uncertainties upon deployment.

Traditionally, domain randomization requires manually defining these sampling distributions, which can be labor-intensive. To address this, recent work has explored methods to automatically learn these distributions, aiming for maximal generalization with minimal manual effort. One notable approach involves constructing adversarial distributions that challenge the current policy, ensuring coverage of the environment parameter space~\cite{mehta2020active, wang2025robustfastadaptationadversarially}. While adversarial methods can outperform uniform randomization, they typically assume all environments sampled are solvable, limiting their effectiveness in highly uncertain scenarios.

An alternative line of research adopts a curriculum-based approach, progressively expanding the complexity of the training distribution while maintaining policy success. Specific curricular methods include minimizing divergence from a target distribution using multivariate Gaussians~\citep{gaussian_dr}, maximizing entropy through independent beta distributions~\citep{entmax}, and incrementally expanding uniform sampling distributions via boundary sampling~\citep{adr}. In this work, we propose a novel learned domain randomization technique leveraging normalizing flows~\cite{rezende2015variational} as neural sampling distributions. This approach offers increased flexibility and expressivity over existing parametric methods.

Beyond training robust policies in simulation, learned sampling distributions can be tied to the real-world environmental conditions under which policies are likely to succeed. Previous works have integrated domain randomization with real-world interactions for more informed training distributions~\citep{bayessim, normflows_adaptive_dr, ajay2023distributionallyadaptivemetareinforcement, bayesian_domain_randomization} or to find the maximally effective real-world strategy ~\cite{strategy_optimization, ren2023adaptsimtaskdrivensimulationadaptation}. However, these methods often necessitate expensive policy retraining or data-intensive evolutionary search based on real-world feedback, posing challenges for real-time applications. Instead, we utilize our learned sampling distribution as an out-of-distribution detector within a multi-step planning framework, enabling fast and data-efficient information gathering in the real world.

\section{Background}

\subsection{Markov Decision Process}
A Markov Decision Process (MDP) is a mathematical framework for modeling decision-making. Formally, an MDP is defined as a tuple $(\mathcal{S}, \mathcal{A}, P, R, \gamma)$, where $\mathcal{S}$ is the state space, $\mathcal{A}$ is the action space, $P: \mathcal{S} \times \mathcal{A} \times \mathcal{S} \rightarrow [0,1]$ is the state transition probability function, where $P(s' \mid s, a)$ denotes the probability of transitioning to state $s'$ from state $s$ after taking action $a$, $R: \mathcal{S} \times \mathcal{A} \rightarrow \mathbb{R}$ is the reward function, where $R(s, a)$ denotes the expected immediate reward received after taking action $a$ in state $s$, $\gamma \in [0,1)$ is the discount factor, representing the importance of future rewards.

A policy $\pi: \mathcal{S} \times \mathcal{A} \rightarrow [0,1]$ defines a probability distribution over actions given states, where $\pi(a \mid s)$ is the probability of taking action $a$ in state $s$. The goal is to find an optimal policy $\pi^*$ that maximizes the expected cumulative discounted reward.

\subsection{Domain Randomization}
Domain randomization introduces variability into the environment by randomizing certain parameters during training. Let $\Xi$ denote the space of domain randomization parameters, and let $\xi \in \Xi$ be a specific instance of these parameters. Each $\xi$ corresponds to a different environment configuration or dynamics.

We can define a parameterized family of Markov Decision Processes (MDPs) where each $\mathcal{M}_\xi = (\mathcal{S}, \mathcal{A}, P_\xi, R_\xi, \gamma)$ has transition dynamics $P_\xi$ and reward function $R_\xi$ dependent on $\xi$. The agent interacts with environments sampled from a distribution over $\Xi$, typically denoted as $p(\xi)$. \footnote{This problem can also be thought of as a POMDP where the observation space is $\mathcal{S}$ and the state space is a product of $\mathcal{S}$ and $\Xi$ as discussed in \cite{latent_mdp}.}

The objective is to learn a policy $\pi: \mathcal{S} \rightarrow \mathcal{A}$ that maximizes the expected return across the distribution environments:

\begin{equation}
J(\pi) = \mathbb{E}_{\xi \sim p(\xi)} \left[ \mathbb{E}_{\tau \sim P_\xi, \pi} \left[ \sum_{t=0}^\infty \gamma^t R_\xi(s_t, a_t) \right] \right],
\end{equation}

where $\tau = \{ (s_0, a_0, s_1, a_1, \dots) \}$ denotes a trajectory generated by policy $\pi$ in environment $\xi$. Domain randomization aims to find a policy $\pi^*$ such that: $\pi^* = \arg\max_\pi J(\pi)$.

In deep reinforcement learning, the policy $\pi$ is a neural network parameterized by $\theta$, denoted as $\pi_\theta$. The agent learns the policy parameters $\theta$ through interactions with simulated environments sampled from $p(\xi)$. In our implementation, we employ the Proximal Policy Optimization (PPO) algorithm \citep{schulman2017proximal}, an on-policy policy gradient method that optimizes a stochastic policy while ensuring stable and efficient learning.

To further stabilize training, we pass privileged information about the environment parameters $\xi$ to the critic network. The critic network, parameterized by $\psi$, estimates the state-value function:

\begin{equation}
V_\psi(s_t, \xi) = \mathbb{E}_{\pi_\theta} \left[ \sum_{k=0}^\infty \gamma^k r_{t+k} \,\bigg|\, s_t, \xi \right],
\end{equation}

where $s_t$ is the current state, $r_{t+k}$ are future rewards, and $\gamma$ is the discount factor. By incorporating $\xi$, the critic can provide more accurate value estimates with lower variance~\citep{pinto2017asymmetric}. The actor network $\pi_\theta(a_t | s_t)$ does not have access to $\xi$, ensuring that the policy relies only on observable aspects of the state.
\begin{figure*}[t!]
    \centering
    \includegraphics[width=0.9\textwidth]{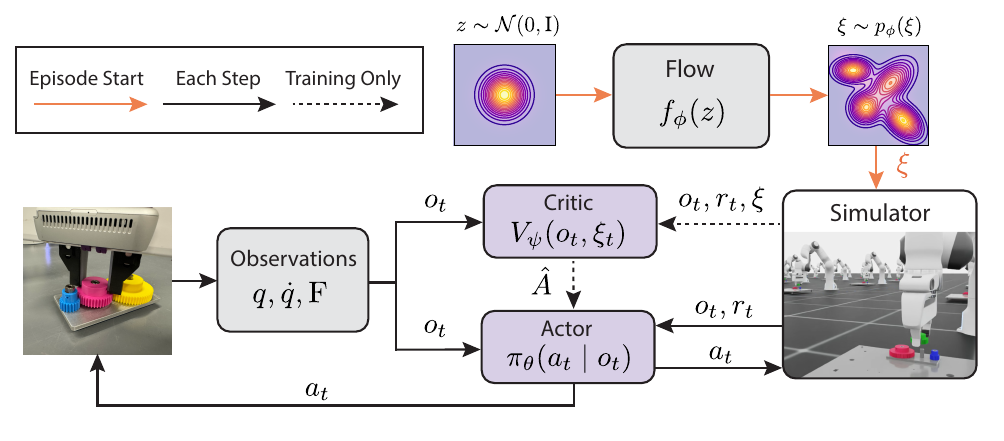}
    \caption{An architecture diagram for our actor-critic RL training setup using a normalizing flow to seed environment parameters across episodes.}
    \label{fig:arch}
\end{figure*}
\subsection{Normalizing Flows}
Normalizing flows are a class of generative models that transform a simple base distribution into a complex target distribution using a sequence of invertible, differentiable functions. Let \( z \sim p_Z(z) \) be a latent variable from a base distribution (e.g., a standard normal distribution). A normalizing flow defines an invertible transformation \( f_{\phi}: \mathbb{R}^d \rightarrow \mathbb{R}^d \) parameterized by neural network parameters \( \phi \), such that \( x = f_{\phi}(z) \), aiming for \( x \) to follow the target distribution.

The density of \( x \) is computed using the change of variables formula:

\begin{equation}
p_X(x) = p_Z(f_{\phi}^{-1}(x)) \left| \det \left( \frac{\partial f_{\phi}^{-1}(x)}{\partial x} \right) \right|.
\end{equation}

For practical computation, this is often rewritten as:

\begin{equation}
\log p_X(x) = \log p_Z(z) - \log \left| \det \left( \frac{\partial f_{\phi}(z)}{\partial z} \right) \right|,
\end{equation}

where \( \frac{\partial f_{\phi}(z)}{\partial z} \) is the Jacobian of \( f_{\phi} \) at \( z \). By composing multiple such transformations \( f_{\phi} = f_{\phi_K} \circ \dots \circ f_{\phi_1} \), each parameterized by neural network parameters \( \phi_k \), normalizing flows can model highly complex distributions.

In our work, we employ \emph{neural spline flows} \citep{durkan2019neural}, a type of normalizing flow where the invertible transformations are constructed using spline-based functions. Specifically, the parameters \( \phi \) represent the coefficients of the splines (e.g., knot positions and heights) and the weights and biases of the neural networks that parameterize these splines. 

\section{Method}

In this section, we introduce \textbf{\OursNoSpace}, a method for learned domain randomization that \emph{goes with the flow} by adaptively adjusting the domain randomization process using normalizing flows.

In traditional domain randomization setups, the distribution \(p(\xi)\) is predefined. 
However, selecting an appropriate \(p(\xi)\) is crucial for the policy's performance and generalization. 
Too broad a sampling distribution and the training focuses on unsolvable environments and falls into local minima. 
In contrast, too narrow a sampling distribution leads to poor generalization and robustness. 
Additionally, rapid changes to the sampling distribution can lead to unstable training.
To address these challenges, prior works such as \cite{selfpaced} have proposed a self-paced learner, which starts by mastering a small set of environments, and gradually expands the tasks to solver harder and harder problems while maintaining training stability. This strategy has subsequently been applied to domain randomization in ~\cite{gaussian_dr} and ~\cite{entmax}, where terms were included for encouraging spread over the sampling space for greater generalization. We take inspiration from these works to form a joint optimization problem:

\begin{equation}
\max_{p, \pi} \left\{ \mathbb{E}_{\xi \sim p} [ J_\xi(\pi) ] + \alpha \mathcal{H}(p) - \beta D_{KL}(p_{\text{old}} \| p) \right\}
\end{equation}

where $p$ is the sampling distribution over environment parameters $p(\xi)$, $\mathcal{H}(p)$ is the differential entropy of $p(\xi)$, \(D_{KL}\left(p \| p_{\text{old}}\right)\) is the divergence between the current and previous sampling distributions, and $\alpha > 0, \beta > 0$ are regularization coefficients that control the trade-off between generalizability, training stability, and the expected reward under the sampling distribution.

Other learned domain randomization approaches propose similar objectives. \cite{gaussian_dr} maximizes reward but replaces entropy regularization with a KL divergence to a fixed target distribution and omits the self-paced KL term. \cite{entmax} includes all three objectives but frames the reward and self-paced KL terms as constraints, maximizing entropy through a nonlinear optimization process that is not easily adaptable to neural sampling distributions. We compare \Ours to these methods in our experiments to highlight its advantages. To our knowledge, \Ours is the first method to optimize such an objective with a neural sampling distribution.

The \Ours algorithm (Algorithm~\ref{alg:goflow}) begins by initializing both the policy parameters $\theta$ and the normalizing flow parameters $\phi$. In each training iteration, \Ours first samples a batch of environment parameters $\{\xi_i\}_{i=1}^B$ from the current distribution modeled by the normalizing flow (Line~\ref{line:sample}). 
These sampled parameters are used to train the policy $\pi_\theta$ (Line~\ref{line:train}). 
Following the policy update, expected returns $J_{\xi_i}(\pi_\theta)$ are estimated for each sampled environment through policy rollouts, providing a measure of the policy's performance under a target uniform distribution $u(\xi)$ after being trained on $p(\xi)$ (Line~\ref{line:rollout}).

\begin{algorithm}[t]
\caption{\OursNoSpace}
\label{alg:goflow}
\begin{algorithmic}[1]
\Require Initial policy parameters $\theta$, flow parameters $\phi$, training steps $N$, network updates $K$, entropy coefficient $\alpha$, similarity coefficient $\beta$, and learning rate $\eta_\phi$
\For{$n = 1$ to $N$}
    \State Sample $\{\xi^{\text{train}}_i\}_{i=1}^B \sim p_\phi(\xi)$, $\{\xi^{\text{test}}_i\}_{i=1}^B \sim u(\xi)$ \label{line:sample}
    \State Train $\pi_\theta$ with $\xi^{\text{train}}_i$ initializations \label{line:train}
    \State Estimate $J_{\xi^{\text{test}}_i}(\pi_\theta)$ via policy rollouts \label{line:rollout}
    
    \State Save current flow distribution as $p_{\phi_{\text{old}}}(\xi)$

    \For{$k=1$ to $K$}
        \State $\mathcal{R} \gets \frac{|\Xi|}{B} \sum_{i=1}^B \Bigl[p_\phi(\xi_i^{\mathrm{test}})\,J_{\xi_i^{\mathrm{test}}}(\pi_\theta)\Bigr].$ \label{line:reward}
        \State $\hat{\mathcal{H}} \gets - |\Xi|\cdot\mathbb{E}_{\xi \sim u(\xi)} \Bigl[ p_\phi(\xi)\log p_\phi(\xi)\Bigr]$ \label{line:entropy}
        \State $\hat{D}_{KL} \gets \mathbb{E}_{\xi \sim p_{\phi_{\text{old}}}(\xi)} \Bigl[ \log p_{\phi_{\text{old}}}(\xi) - \log p_\phi(\xi) \Bigr]$ \label{line:kl}
        \State $\phi \leftarrow \phi + \eta_\phi \nabla_\phi \left(\mathcal{R} + \alpha \hat{H} - \beta \hat{D}_{KL} \right)$ \label{line:gradient}
    \EndFor
\EndFor
\end{algorithmic}
\end{algorithm}

\begin{figure*}[t!]
    \centering
    \includegraphics[width=\textwidth]{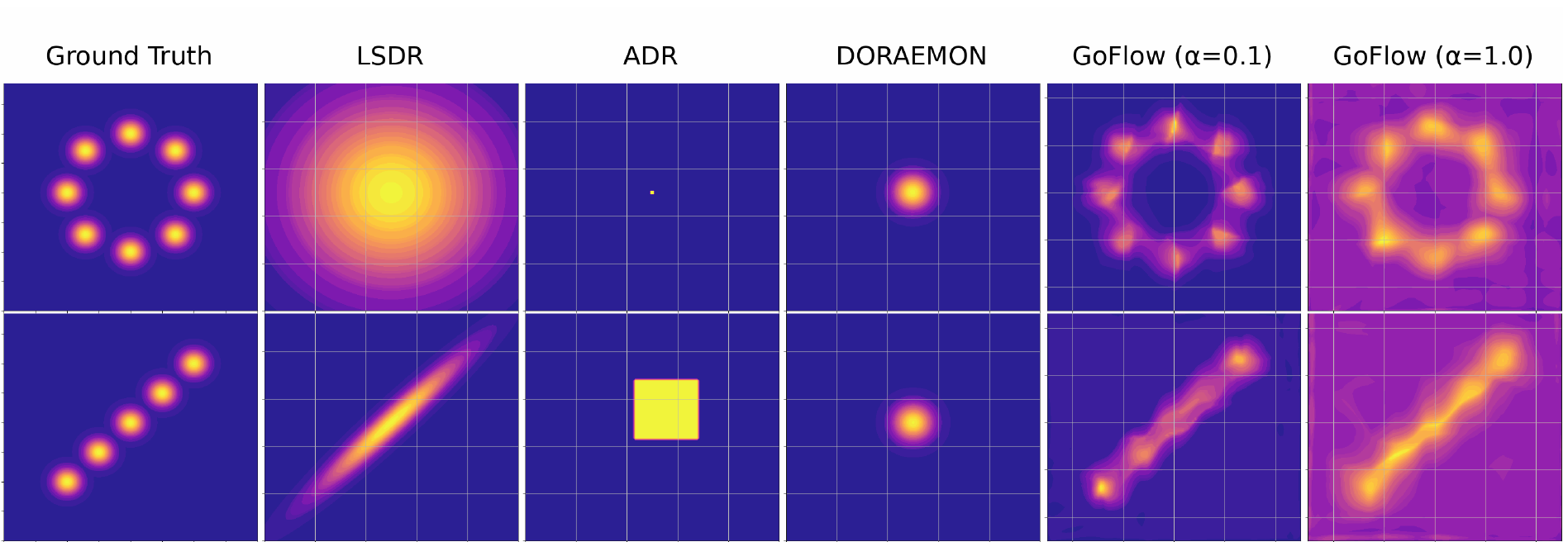}
    \caption{An illustrative domain showing the learned sampling functions over the space of unobserved parameters for the tested baselines. Compared to other learning methods, \Ours correctly models the multimodality and inter-variable dependencies of the underlying reward function. This toy domain, along with other domains in our experiments, violates some of the assumptions made by prior works, such as the feasibility of the center point of the range.}
    \label{fig:toy_sampling_dist}
\end{figure*}

After sampling these rollouts, \Ours then performs $K$ steps of optimization on the sampling distribution.
\Ours first estimates the policy's performance on the sampling distribution using the previously sampled rollout trajectories (Line~\ref{line:reward}).
The entropy of the sampling distribution (Line~\ref{line:entropy}) and divergence from the previous sampling distribution (Line~\ref{line:kl}) are estimated using newly drawn samples. Importantly, we compute the reward and entropy terms by importance sampling from a uniform distribution rather than from the flow itself. This broad coverage helps prevent the learned distribution from collapsing around a small region of parameter space. Derivations for these equations can be found in Appendix~\ref{importance_sampling_proofs}. These terms are combined to form a loss, which is differentiated to update the parameters of the sampling distribution (Line~\ref{line:gradient}). An architecture diagram for this approach can be seen in Figure~\ref{fig:arch}.

\section{Domain Randomization Experiments}

Our simulated experiments compare policy robustness in a range of domains. For full details on the randomization parameters and bounds, see Appendix~\ref{app:dr_params}.
\begin{figure*}[ht!]
    \centering
    \includegraphics[width=\linewidth]{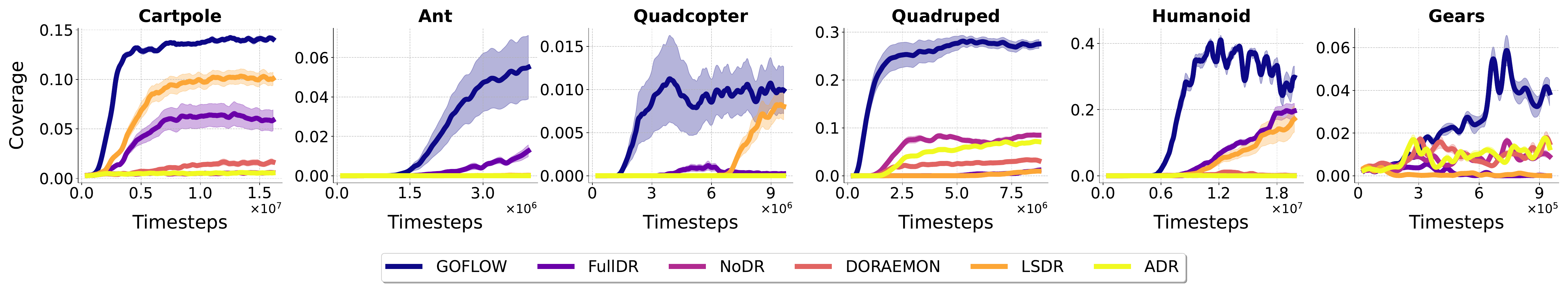}
    \caption{The coverage ratio over the target distribution across five random seeds for each of the environments. The bands around each curve indicate the standard error.}
    \label{fig:sim_results}
\end{figure*}
\subsection{Domains}
First, we examine the application of \Ours to an illustrative 2D domain that is multimodal and contains intervariable dependencies. The state and action space are in $\mathbb{R}^2$. The agent is initialized randomly in a bounded x, y plane. An energy function is defined by a composition of 
Gaussians placed in a regular circular or linear array. The agent can observe its position with Gaussian noise proportional to the inverse of the energy function. The agent is rewarded for guessing its location, but is incapable of moving. This task is infeasible when the agent is sufficiently far from any of the 
Gaussian centers, so a sampling distribution should come to resemble the energy function. Some example functions learned by \Ours and baselines from Section~\ref{sec:baselines} can be seen in Figure~\ref{fig:toy_sampling_dist}.

Second, we quantitatively compare \Ours to existing baselines including Cartpole, Ant, Quadcopter, Quadruped, and Humanoid in the IsaacLab suite of environments~\citep{isaaclab}. We randomize over parameters such as link masses, joint frictions, and material properties.

Lastly, we evaluate our method on a contact-rich robot manipulation task of gear insertion, a particularly relevant problem for robotic assembly. 
In the gears domain, we randomize over the relative pose between the gripper and the held gears along three degrees of freedom. 
The problem is made difficult by the uncertainty the robot has about the precise location of the gear relative to the hand.
The agent must learn to rely on signals of proprioception and force feedback to guide the gear into the gear shaft. 
The action space consists of end-effector pose offsets along three translational degrees of freedom and one rotational degree around the z dimension. 
The observation space consists of a history of the ten previous end effector poses and velocities estimated via finite differencing.
In addition to simulated experiments, our trained policies are tested on a Franka Emika robot using \texttt{IndustReal} library built on the \texttt{frankapy} toolkit~\citep{tang2023industreal, zhang2020modular}.

\subsection{Baselines}
\label{sec:baselines}

In our domain randomization experiments, we compare to a number of standard RL baselines and learning-based approaches from the literature. In our quantitative experiments, success is determined by the sampled environment passing a certain performance threshold $J_T$ that was selected for each environment. We measure task performance in terms of \textit{coverage}, which is defined as the proportion of the total sampling distribution for which the policy receives higher than $J_t$ reward. Coverage is estimated via policy rollouts on 4096 uniformly selected environment initialization samples from the total range of parameters. All baselines are trained with an identical neural architecture and PPO implementation. The success thresholds along with other hyperparameters are in Appendix~\ref{app:hyperparameters}.

We evaluate on the following baselines. First, we compare to no domain randomization (\textbf{NoDR}) which trains on a fixed environment parameter at the centroid of the parameter space. Next, we compare to a full domain randomization (\textbf{FullDR}) which samples uniformly across the domain within the boundaries during training. In addition to these fixed randomization methods, we evaluate against some other learning-based solutions from the literature: \textbf{ADR}~\citep{adr} learns uniform intervals that expand over time via boundary sampling. It starts by occupying an initial percentage of the domain and performs ``boundary sampling'' during training with some probability. The rewards attained from boundary sampling are compared to thresholds that determine if the boundary should be expanded or contracted.  \textbf{LSDR}~\citep{gaussian_dr} learns a multivariate gaussian sampling distribution using reward maximization with a KL divergence regularization term weighted by an $\alpha$ hyperparameter. Lastly, \textbf{DORAEMON}~\citep{entmax} learns independent beta distributions for each dimension of the domain, using a maximum entropy objective constrained by an estimated success rate.

We compare coverage across environments sampled from a uniform testing distribution within the environment bounds. Our findings in Figure~\ref{fig:sim_results} show that \Ours matches or outperforms baselines across all domains. We find that our method performs particularly well in comparison to other learned baselines when the simpler or more feasible regions of the domain are off-center, irregularly shaped, and have inter-parameter dependencies such as those seen in Figure~\ref{fig:toy_sampling_dist}. We intentionally chose large intervals with low coverage to demonstrate this capability, and we perform a full study of how baselines degrade significantly with increased parameter ranges while \Ours degrades more gracefully (see Appendix~\ref{app:coverage_experiments}). Additionally, coverage should not be mistaken for success-rate, as these policies can be integrated into a planning framework that avoids the use of infeasible actions or gathers information to increase coverage as discussed in the following sections.

In addition to simulated experiments, we additionally evaluated the trained policies on a real-world gear insertion task. The results of those real-world experiments show that \Ours results in more robust sim-to-real transfer as seen in Table~\ref{tab:real_world} and in the supplementary videos.

\begin{figure*}[ht!]
    \centering
    \includegraphics[width=\textwidth]{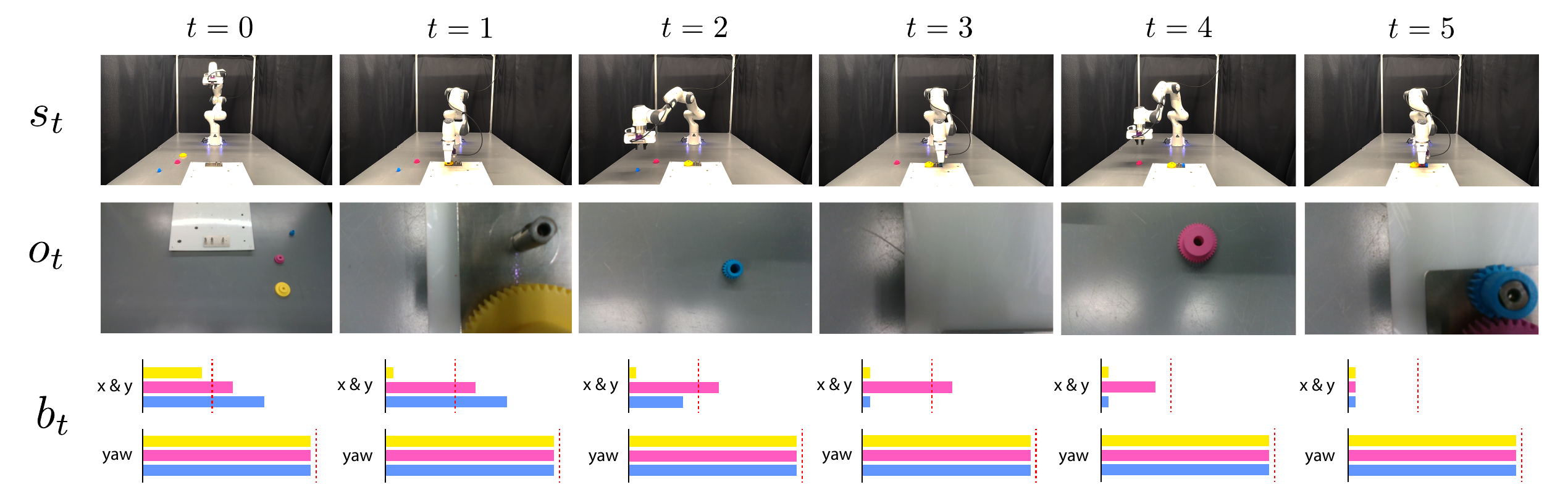}
    \caption{A multi-step manipulation plan using probabilistic pose estimation to estimate and update beliefs over time. The three rows show the robot state $s_t$, the observation $o_t$, and the robot belief $b_t$ at each timestep. The red dotted line in the belief indicates the marginal entropy thresholds for the x, y, and yaw (rotation around z) dimensions as determined by the learned normalizing flow. A belief with entropy surpassing the threshold line indicates the policy will likely fail. For full visualizations of the belief posteriors, flow distributions, and value maps, see Figure~\ref{fig:belief_posteriors}.}
    \label{fig:multi-step-plan}
\end{figure*}

\section{Application to Multi-step manipulation}

While reinforcement learning has proven to be a valuable technique for learning short-horizon skills in dynamic and contact-rich settings, it often struggles to generalize to more long-horizon and open ended problems~\citep{rl_is_hard}. 
The topic of sequencing short horizon skills in the context of a higher-level decision strategy has been of increasing interest to both the planning~\citep{mishra2023generativeskillchaininglonghorizon} and reinforcement learning communities~\citep{maple}.
For this reason, we examine the utility of these learned sampling distributions as out-of-distribution detectors, or belief-space preconditions, in the context of a multi-step planning system.

\subsection{Belief-space planning background}

Belief-space planning is a framework for decision-making under uncertainty, where the agent maintains a probability distribution over possible states, known as the \emph{belief state}. Instead of planning solely in the state space \(\mathcal{S}\), the agent operates in the \emph{belief space} \(\mathcal{B}\), which consists of all possible probability distributions over \(\mathcal{S}\). This approach is particularly useful in partially observable environments where there is uncertainty in environment parameters and where it is important to take actions to gain information.

Rather than operating at the primitive action level, belief-space planners often make use of high-level actions $\mathcal{A}_\Pi$, sometimes called skills or options. 
In our case, these high-level actions will be a discrete set of pretrained RL policies. 
These high-level actions come with a \emph{belief-space precondition} and a \emph{belief-space effect}, both of which are subsets of the belief space \(\mathcal{B}\) ~\cite{BHPN, tampura}. 
Specifically, a high-level action \( \pi \in \mathcal{A}_\Pi \) is associated with two components: a precondition \( \text{Pre}_\pi \subseteq \mathcal{B} \), representing the set of belief states from which the action can be applied, and an effect \( \text{Eff}_\pi \subseteq \mathcal{B} \), representing the set of belief states that the action was designed or trained to achieve. 
If the precondition holds—that is, if the current belief state \( b \) satisfies \( b \in \text{Pre}_\pi \)—then applying action \( a \) will achieve the effect \( \text{Eff}_\pi \) with probability at least \( \eta \in [0,1] \). 
More formally, if $b \in \text{Pre}_\pi$, then $\Pr\left( b_{t+1} \in \text{Eff}_\pi \,\big|\, b_t, \pi \right) \geq \eta$. Depending on the planner, $\eta$ may be set in advance, or calculated by the planner as a function of the belief.

This formalization allows planners to reason abstractly about the effects of high-level actions under uncertainty, which can result in generalizable decision-making in long-horizon problems that require active information gathering or risk-awareness.

\subsection{Computing preconditions}
\label{sec:computing_preconditions}
In this section, we highlight the potential application of the learned sampling distribution $p_\phi$ and privileged value function $V_\psi$ as a useful artifacts for belief-space planning. In particular, we are interested in identifying belief-space preconditions of a set of trained skills.

One point of leverage we have for this problem is the privileged value function $V_\psi(s, \xi)$, which was learned alongside the policy during training. 
One way to estimate the belief-space precondition is to simply find the set of belief states for which the expected value of the policy is larger than $J_T$ with probability greater than $\eta$ under the belief: 
\begin{equation}
\text{Pre}_\pi = \left\{ b \in \mathcal{B} \mid \mathbb{E}_{b}\left[\mathbbm{1}_{V_\psi(s, \xi)> J_T}\right] > \eta  \right\}.
\end{equation}

\begin{figure*}[ht!]
    \centering
    \includegraphics[width=0.95\textwidth]{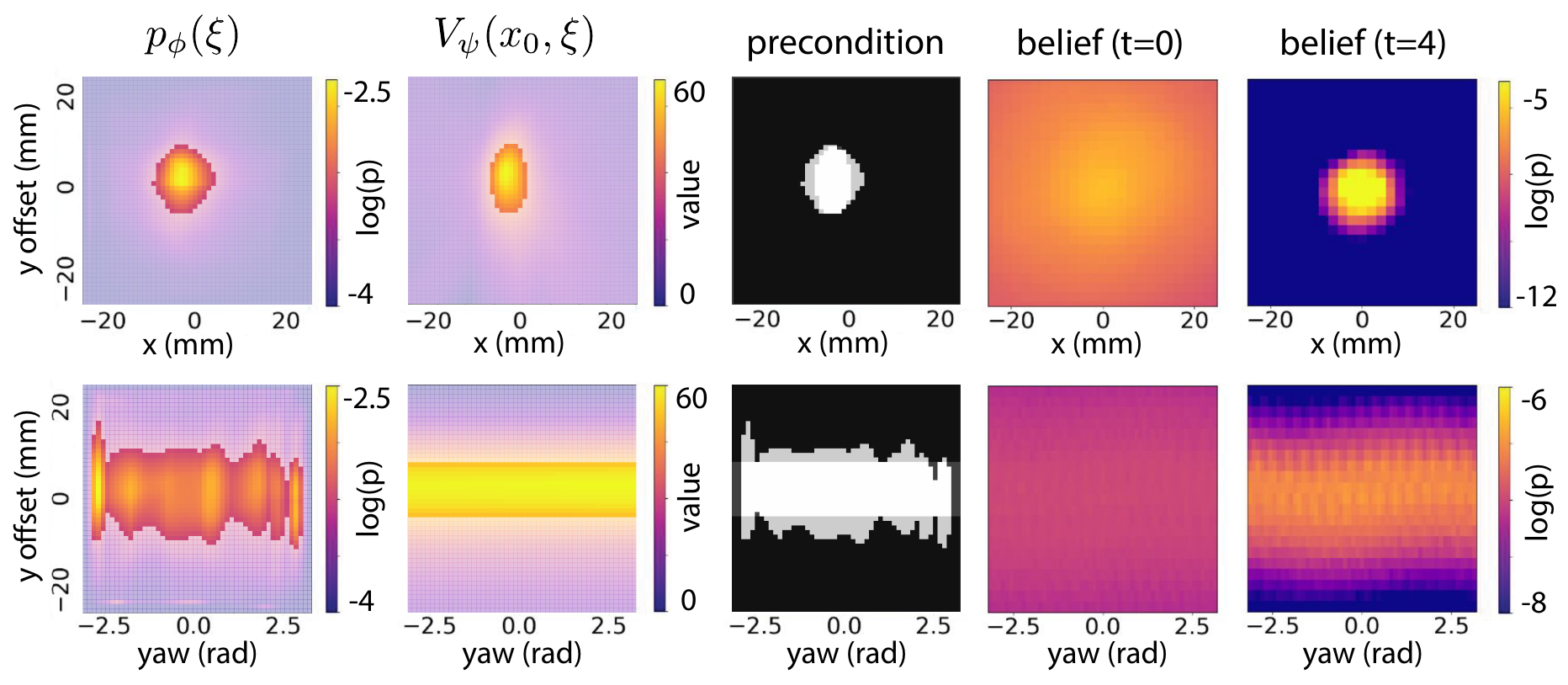}
    \caption{A visual example of the precondition computation described in Section~\ref{sec:computing_preconditions} for the gear assembly plan shown in Figure~\ref{fig:multi-step-plan}. The two rows show two different projections of the 3D sampling space (x position vs y position in the top row and y position vs yaw rotation in the bottom row). We apply a threshold $\epsilon$ to the sampling distribution to remove low-probability regions (column 1). Additionally, we filter the value function by retaining only the regions where the expected value exceeds a predetermined threshold $\eta$ (column 2). The intersection of these two regions defines the belief-space precondition, indicating where the policy is likely to succeed (column 3). 
    Comparing the precondition to the beliefs, we can see that the belief is not sufficiently contained within the precondition at $t=0$ (column 4), but passes the success threshold $\eta$ at after closer inspection at $t=4$ (column 5).}
    \label{fig:planning_dists}
\end{figure*}

However, a practical issue with this computation is that the value function is likely not calibrated in large portions of the state space that were not seen during policy training.
To address this, we focus on regions of the environment where the agent has higher confidence due to sufficient sampling, i.e., where $p_\phi(\xi) > \epsilon$ for a threshold $\epsilon$. This enables us to integrate the value function over the belief distribution $b(x, \xi)$ and the trusted region within $\Xi$:

\begin{equation}
\text{Pre}_\pi = \left\{ b \in \mathcal{B} \mid \mathbb{E}_{b}\left[\mathbbm{1}_{V_\psi(s, \xi)> J_T} \cdot \mathbbm{1}_{p_\phi(\xi) > \epsilon}\right] > \eta \right\}
\end{equation}.

Here, $\eta$ lower bounds the probability of achieving the desired effects $\text{Eff}_\pi$ (or value greater than $J_T$) after executing $\pi$ in any belief state in $\text{Pre}_\pi$.
Figure~\ref{fig:planning_dists} shows an example precondition for a single step of the assembly plan.

\subsection{Updating beliefs}
\label{sec:belief_update}

Updating the belief state requires a probabilistic state estimation system that outputs a posterior over the unobserved environment variables, rather than a single point estimate. 
We use a probabilistic object pose estimation framework called Bayes3D to infer posterior distributions over object pose~\citep{gothoskar2023bayes3d}. 
For details on this, see Appendix~\ref{app:belief_update}.

The benefit of this approach in contrast to traditional rendering-based pose estimation systems, such as those presented in  \cite{wen2024foundationposeunified6dpose} or \cite{megapose}, is that pose estimates from Bayes3D indicate high uncertainty for distant, small, or occluded objects as well as uncertainty stemming from object symmetry. Figure~\ref{fig:belief_posteriors} shows the pose beliefs across the multi-step plan.

\subsection{A simple belief-space planner}
While the problem of general-purpose multi-step planning in belief-space has been widely studied, in this paper we use a simple BFS belief-space planner to demonstrate the utility of the learned sampling distributions as belief-space preconditions. 
The full algorithm can be found in Algorithm~\ref{alg:belief_space_planner}. 

An example plan can be seen in Figure~\ref{fig:multi-step-plan}. 
The goal is to assemble the gear box by inserting all three gears (yellow, pink, and blue) into the shafts on the gear plate. 
Each gear insertion is associated with a separate policy for each color trained with \OursNoSpace. 
In addition to the trained policies, the robot is given access to an object-parameterized inspection action which has no preconditions and whose effects are a reduced-variance pose estimate attained by moving the camera closer to the object. 
The robot is initially uncertain of the x, y, and yaw components of the 6-dof pose based on probabilistic pose estimates. Despite this uncertainty, the robot is confident enough in the pose of the largest and closest yellow gear to pick it up and insert it. 
In contrast, the blue and pink gears require further inspection to get a better pose estimate.
Closer inspection reduces uncertainty along the x and y axis, but reveals no additional information about yaw dimension due to rotational symmetry. 
Despite an unknown yaw dimension, the robot is confident in the insertion because the flow $p_\phi$ indicates that success is invariant to the yaw dimension. This is due to the fact that success in the insertion task is defined by the distance between the bottom center of the gear and the base of the gear shaft, which is independent of gear rotation.
For visualizations of the beliefs and flows at each step, see Appendix~\ref{app:multi-step-planning}.

\section{Conclusion and discussion}

In this paper, we introduced \OursNoSpace, a novel approach to domain randomization that uses normalizing flows to dynamically adjust the sampling distribution during reinforcement learning. 
By combining actor-critic reinforcement learning with a learned neural sampling distribution, we enabled more flexible and expressive parameterization of environmental variables, leading to better generalization in complex tasks like contact-rich assembly. 
Our experiments demonstrated that GoFlow achieves higher coverage than traditional fixed and learning-based domain randomization techniques across a variety of simulated environments, particularly in scenarios where the domain has irregular dependencies between parameters. 
The method also showed promise in real-world robotic tasks including contact-rich assembly.

Moreover, we extended GoFlow to multi-step decision-making tasks, integrating it with belief-space planning to handle long-horizon problems under uncertainty. 
This extension enabled the use of learned sampling distributions and value functions as preconditions leading to active information gathering. 

Although \Ours enables more expressive sampling distributions, it also presents some new challenges. One limitation of our method is that it has higher variance due to occasional training instability of the flow. 
This instability can be alleviated by increasing $\beta$, but at the cost of reduced sample efficiency (see Appendix~\ref{app:hyperparameters}). 
In addition, using the flow and value estimates for belief-space planning require manual selection and tuning of several thresholds which are environment specific. 
The $\eta$ parameter may be converted from a threshold into a cost in the belief space planner, which would remove one point of manual tuning. 
However, removing the $\epsilon$ parameter may prove more difficult, as it would require uncertainty quantification of the neural value function. 
Despite these challenges, we hope this work inspires further research on integrating short-horizon learned policies into broader planning frameworks, particularly in contexts involving uncertainty and partial observability.

\section{Impact Statement}
This paper presents work whose goal is to advance the field of Machine Learning. There are many potential societal consequences of our work, none which we feel must be specifically highlighted here.

\bibliography{example_paper}

\begin{thebibliography}{49}
\providecommand{\natexlab}[1]{#1}
\providecommand{\url}[1]{\texttt{#1}}
\expandafter\ifx\csname urlstyle\endcsname\relax
  \providecommand{\doi}[1]{doi: #1}\else
  \providecommand{\doi}{doi: \begingroup \urlstyle{rm}\Url}\fi

\bibitem[Ajay et~al.(2023)Ajay, Gupta, Ghosh, Levine, and Agrawal]{ajay2023distributionallyadaptivemetareinforcement}
Ajay, A., Gupta, A., Ghosh, D., Levine, S., and Agrawal, P.
\newblock Distributionally adaptive meta reinforcement learning, 2023.
\newblock URL \url{https://arxiv.org/abs/2210.03104}.

\bibitem[Chebotar et~al.(2019)Chebotar, Handa, Makoviychuk, Macklin, Issac, Ratliff, and Fox]{chebotar2019closing}
Chebotar, Y., Handa, A., Makoviychuk, V., Macklin, M., Issac, J., Ratliff, N., and Fox, D.
\newblock Closing the sim-to-real loop: Adapting simulation randomization with real world experience.
\newblock In \emph{2019 International Conference on Robotics and Automation (ICRA)}, pp.\  8973--8979. IEEE, 2019.

\bibitem[Chen et~al.(2021)Chen, Hu, Jin, Li, and Wang]{understanding_dr}
Chen, X., Hu, J., Jin, C., Li, L., and Wang, L.
\newblock Understanding domain randomization for sim-to-real transfer.
\newblock \emph{CoRR}, abs/2110.03239, 2021.
\newblock URL \url{https://arxiv.org/abs/2110.03239}.

\bibitem[Curtis et~al.(2022)Curtis, Kaelbling, and Jain]{curtis2022taskdirectedexplorationcontinuouspomdps}
Curtis, A., Kaelbling, L., and Jain, S.
\newblock Task-directed exploration in continuous pomdps for robotic manipulation of articulated objects, 2022.
\newblock URL \url{https://arxiv.org/abs/2212.04554}.

\bibitem[Curtis et~al.(2024)Curtis, Matheos, Gothoskar, Mansinghka, Tenenbaum, Lozano-Pérez, and Kaelbling]{tampura}
Curtis, A., Matheos, G., Gothoskar, N., Mansinghka, V., Tenenbaum, J., Lozano-Pérez, T., and Kaelbling, L.~P.
\newblock Partially observable task and motion planning with uncertainty and risk awareness, 2024.
\newblock URL \url{https://arxiv.org/abs/2403.10454}.

\bibitem[Del~Moral et~al.(2006)Del~Moral, Doucet, and Jasra]{del2006sequential}
Del~Moral, P., Doucet, A., and Jasra, A.
\newblock Sequential monte carlo samplers.
\newblock \emph{Journal of the Royal Statistical Society Series B: Statistical Methodology}, 68\penalty0 (3):\penalty0 411--436, 2006.

\bibitem[Durkan et~al.(2019)Durkan, Bekasov, Murray, and Papamakarios]{durkan2019neural}
Durkan, C., Bekasov, A., Murray, I., and Papamakarios, G.
\newblock Neural spline flows.
\newblock In \emph{Advances in Neural Information Processing Systems (NeurIPS)}, volume~32, 2019.

\bibitem[Gothoskar et~al.(2023)Gothoskar, Ghavami, Li, Curtis, Noseworthy, Chung, Patton, Freeman, Tenenbaum, Klukas, et~al.]{gothoskar2023bayes3d}
Gothoskar, N., Ghavami, M., Li, E., Curtis, A., Noseworthy, M., Chung, K., Patton, B., Freeman, W.~T., Tenenbaum, J.~B., Klukas, M., et~al.
\newblock Bayes3d: fast learning and inference in structured generative models of 3d objects and scenes.
\newblock \emph{arXiv preprint arXiv:2312.08715}, 2023.

\bibitem[Haarnoja et~al.(2018)Haarnoja, Zhou, Abbeel, and Levine]{haarnoja2018soft}
Haarnoja, T., Zhou, A., Abbeel, P., and Levine, S.
\newblock Soft actor-critic: Off-policy maximum entropy deep reinforcement learning with a stochastic actor.
\newblock In \emph{International conference on machine learning}, pp.\  1861--1870. PMLR, 2018.

\bibitem[Jin et~al.(2023)Jin, Lin, Song, Li, and Yang]{Jin2023}
Jin, P., Lin, Y., Song, Y., Li, T., and Yang, W.
\newblock Vision-force-fused curriculum learning for robotic contact-rich assembly tasks.
\newblock \emph{Frontiers in Neurorobotics}, 17:\penalty0 1280773, October 2023.
\newblock \doi{10.3389/fnbot.2023.1280773}.

\bibitem[Josifovski et~al.(2022)Josifovski, Malmir, Klarmann, {\v{Z}}agar, Navarro-Guerrero, and Knoll]{josifovski2022analysis}
Josifovski, J., Malmir, M., Klarmann, N., {\v{Z}}agar, B.~L., Navarro-Guerrero, N., and Knoll, A.
\newblock Analysis of randomization effects on sim2real transfer in reinforcement learning for robotic manipulation tasks.
\newblock In \emph{2022 IEEE/RSJ International Conference on Intelligent Robots and Systems (IROS)}, pp.\  10193--10200. IEEE, 2022.

\bibitem[Kaelbling \& Lozano-Perez(2013)Kaelbling and Lozano-Perez]{BHPN}
Kaelbling, L. and Lozano-Perez, T.
\newblock Integrated task and motion planning in belief space.
\newblock \emph{The International Journal of Robotics Research}, 32:\penalty0 1194--1227, 08 2013.
\newblock \doi{10.1177/0278364913484072}.

\bibitem[Klink et~al.(2021)Klink, Abdulsamad, Belousov, D'Eramo, Peters, and Pajarinen]{selfpaced}
Klink, P., Abdulsamad, H., Belousov, B., D'Eramo, C., Peters, J., and Pajarinen, J.
\newblock A probabilistic interpretation of self-paced learning with applications to reinforcement learning.
\newblock \emph{CoRR}, abs/2102.13176, 2021.
\newblock URL \url{https://arxiv.org/abs/2102.13176}.

\bibitem[Kober et~al.(2013)Kober, Bagnell, and Peters]{kober2013reinforcement}
Kober, J., Bagnell, J.~A., and Peters, J.
\newblock Reinforcement learning in robotics: A survey.
\newblock \emph{The International Journal of Robotics Research}, 32\penalty0 (11):\penalty0 1238--1274, 2013.

\bibitem[Kwon et~al.(2021)Kwon, Efroni, Caramanis, and Mannor]{latent_mdp}
Kwon, J., Efroni, Y., Caramanis, C., and Mannor, S.
\newblock {RL} for latent mdps: Regret guarantees and a lower bound.
\newblock \emph{CoRR}, abs/2102.04939, 2021.
\newblock URL \url{https://arxiv.org/abs/2102.04939}.

\bibitem[Labbé et~al.(2022)Labbé, Manuelli, Mousavian, Tyree, Birchfield, Tremblay, Carpentier, Aubry, Fox, and Sivic]{megapose}
Labbé, Y., Manuelli, L., Mousavian, A., Tyree, S., Birchfield, S., Tremblay, J., Carpentier, J., Aubry, M., Fox, D., and Sivic, J.
\newblock Megapose: 6d pose estimation of novel objects via render \& compare, 2022.
\newblock URL \url{https://arxiv.org/abs/2212.06870}.

\bibitem[Liang et~al.(2020)Liang, Saxena, and Kroemer]{task_oriented_exploration}
Liang, J., Saxena, S., and Kroemer, O.
\newblock Learning active task-oriented exploration policies for bridging the sim-to-real gap.
\newblock \emph{CoRR}, abs/2006.01952, 2020.
\newblock URL \url{https://arxiv.org/abs/2006.01952}.

\bibitem[Luo \& Li(2021)Luo and Li]{luo2021learning}
Luo, J. and Li, H.
\newblock A learning approach to robot-agnostic force-guided high precision assembly.
\newblock In \emph{2021 IEEE/RSJ International Conference on Intelligent Robots and Systems (IROS)}, pp.\  2151--2157. IEEE, 2021.

\bibitem[Mehta et~al.(2020)Mehta, Diaz, Golemo, Pal, and Paull]{mehta2020active}
Mehta, B., Diaz, M., Golemo, F., Pal, C.~J., and Paull, L.
\newblock Active domain randomization.
\newblock In \emph{Conference on Robot Learning}, pp.\  1162--1176. PMLR, 2020.

\bibitem[Mishra et~al.(2023)Mishra, Xue, Chen, and Xu]{mishra2023generativeskillchaininglonghorizon}
Mishra, U.~A., Xue, S., Chen, Y., and Xu, D.
\newblock Generative skill chaining: Long-horizon skill planning with diffusion models, 2023.
\newblock URL \url{https://arxiv.org/abs/2401.03360}.

\bibitem[Mittal et~al.(2023)Mittal, Yu, Yu, Liu, Rudin, Hoeller, Yuan, Singh, Guo, Mazhar, Mandlekar, Babich, State, Hutter, and Garg]{isaaclab}
Mittal, M., Yu, C., Yu, Q., Liu, J., Rudin, N., Hoeller, D., Yuan, J.~L., Singh, R., Guo, Y., Mazhar, H., Mandlekar, A., Babich, B., State, G., Hutter, M., and Garg, A.
\newblock Orbit: A unified simulation framework for interactive robot learning environments.
\newblock \emph{IEEE Robotics and Automation Letters}, 8\penalty0 (6):\penalty0 3740--3747, 2023.
\newblock \doi{10.1109/LRA.2023.3270034}.

\bibitem[Mozifian et~al.(2019)Mozifian, Higuera, Meger, and Dudek]{gaussian_dr}
Mozifian, M., Higuera, J. C.~G., Meger, D., and Dudek, G.
\newblock Learning domain randomization distributions for transfer of locomotion policies.
\newblock \emph{CoRR}, abs/1906.00410, 2019.
\newblock URL \url{http://arxiv.org/abs/1906.00410}.

\bibitem[Muratore et~al.(2019)Muratore, Gienger, and Peters]{muratore2019assessing}
Muratore, F., Gienger, M., and Peters, J.
\newblock Assessing transferability from simulation to reality for reinforcement learning.
\newblock \emph{IEEE transactions on pattern analysis and machine intelligence}, 43\penalty0 (4):\penalty0 1172--1183, 2019.

\bibitem[Muratore et~al.(2020)Muratore, Eilers, Gienger, and Peters]{bayesian_domain_randomization}
Muratore, F., Eilers, C., Gienger, M., and Peters, J.
\newblock Bayesian domain randomization for sim-to-real transfer.
\newblock \emph{CoRR}, abs/2003.02471, 2020.
\newblock URL \url{https://arxiv.org/abs/2003.02471}.

\bibitem[Muratore et~al.(2022)Muratore, Gruner, Wiese, Belousov, Gienger, and Peters]{pmlr-v164-muratore22a}
Muratore, F., Gruner, T., Wiese, F., Belousov, B., Gienger, M., and Peters, J.
\newblock Neural posterior domain randomization.
\newblock In Faust, A., Hsu, D., and Neumann, G. (eds.), \emph{Proceedings of the 5th Conference on Robot Learning}, volume 164 of \emph{Proceedings of Machine Learning Research}, pp.\  1532--1542. PMLR, 08--11 Nov 2022.
\newblock URL \url{https://proceedings.mlr.press/v164/muratore22a.html}.

\bibitem[Nasiriany et~al.(2021)Nasiriany, Liu, and Zhu]{maple}
Nasiriany, S., Liu, H., and Zhu, Y.
\newblock Augmenting reinforcement learning with behavior primitives for diverse manipulation tasks.
\newblock \emph{CoRR}, abs/2110.03655, 2021.
\newblock URL \url{https://arxiv.org/abs/2110.03655}.

\bibitem[Noseworthy et~al.(2024)Noseworthy, Tang, Wen, Handa, Roy, Fox, Ramos, Narang, and Akinola]{forge}
Noseworthy, M., Tang, B., Wen, B., Handa, A., Roy, N., Fox, D., Ramos, F., Narang, Y., and Akinola, I.
\newblock Forge: Force-guided exploration for robust contact-rich manipulation under uncertainty, 2024.
\newblock URL \url{https://arxiv.org/abs/2408.04587}.

\bibitem[OpenAI et~al.(2019)OpenAI, Akkaya, Andrychowicz, Chociej, Litwin, McGrew, Petron, Paino, Plappert, Powell, Ribas, Schneider, Tezak, Tworek, Welinder, Weng, Yuan, Zaremba, and Zhang]{adr}
OpenAI, Akkaya, I., Andrychowicz, M., Chociej, M., Litwin, M., McGrew, B., Petron, A., Paino, A., Plappert, M., Powell, G., Ribas, R., Schneider, J., Tezak, N., Tworek, J., Welinder, P., Weng, L., Yuan, Q., Zaremba, W., and Zhang, L.
\newblock Solving rubik's cube with a robot hand.
\newblock \emph{CoRR}, abs/1910.07113, 2019.
\newblock URL \url{http://arxiv.org/abs/1910.07113}.

\bibitem[Packer et~al.(2018)Packer, Gao, Kos, Kr{\"{a}}henb{\"{u}}hl, Koltun, and Song]{DBLP:journals/corr/abs-1810-12282}
Packer, C., Gao, K., Kos, J., Kr{\"{a}}henb{\"{u}}hl, P., Koltun, V., and Song, D.
\newblock Assessing generalization in deep reinforcement learning.
\newblock \emph{CoRR}, abs/1810.12282, 2018.
\newblock URL \url{http://arxiv.org/abs/1810.12282}.

\bibitem[Peng et~al.(2017)Peng, Andrychowicz, Zaremba, and Abbeel]{original_dr}
Peng, X.~B., Andrychowicz, M., Zaremba, W., and Abbeel, P.
\newblock Sim-to-real transfer of robotic control with dynamics randomization.
\newblock \emph{CoRR}, abs/1710.06537, 2017.
\newblock URL \url{http://arxiv.org/abs/1710.06537}.

\bibitem[Pinto et~al.(2017)Pinto, Andrychowicz, Welinder, Zaremba, and Abbeel]{pinto2017asymmetric}
Pinto, L., Andrychowicz, M., Welinder, P., Zaremba, W., and Abbeel, P.
\newblock Asymmetric actor critic for image-based robot learning.
\newblock \emph{CoRR}, abs/1710.06542, 2017.
\newblock URL \url{http://arxiv.org/abs/1710.06542}.

\bibitem[Ramos et~al.(2019)Ramos, Possas, and Fox]{bayessim}
Ramos, F., Possas, R.~C., and Fox, D.
\newblock Bayessim: adaptive domain randomization via probabilistic inference for robotics simulators.
\newblock \emph{CoRR}, abs/1906.01728, 2019.
\newblock URL \url{http://arxiv.org/abs/1906.01728}.

\bibitem[Ren et~al.(2023)Ren, Dai, Burchfiel, and Majumdar]{ren2023adaptsimtaskdrivensimulationadaptation}
Ren, A.~Z., Dai, H., Burchfiel, B., and Majumdar, A.
\newblock Adaptsim: Task-driven simulation adaptation for sim-to-real transfer, 2023.
\newblock URL \url{https://arxiv.org/abs/2302.04903}.

\bibitem[Rezende \& Mohamed(2015)Rezende and Mohamed]{rezende2015variational}
Rezende, D.~J. and Mohamed, S.
\newblock Variational inference with normalizing flows.
\newblock In \emph{Proceedings of the 32nd International Conference on Machine Learning (ICML)}. PMLR, 2015.

\bibitem[Rozet et~al.(2022)]{rozet2022zuko}
Rozet, F. et~al.
\newblock {Zuko}: Normalizing flows in pytorch, 2022.
\newblock URL \url{https://pypi.org/project/zuko}.

\bibitem[Sagawa \& Hino(2024)Sagawa and Hino]{normflows_adaptive_dr}
Sagawa, S. and Hino, H.
\newblock Gradual domain adaptation via normalizing flows, 2024.
\newblock URL \url{https://arxiv.org/abs/2206.11492}.

\bibitem[Schoettler et~al.(2020)Schoettler, Nair, Luo, Bahl, Ojea, Solowjow, and Levine]{schoettler2020deep}
Schoettler, G., Nair, A., Luo, J., Bahl, S., Ojea, J.~A., Solowjow, E., and Levine, S.
\newblock Deep reinforcement learning for industrial insertion tasks with visual inputs and natural rewards.
\newblock In \emph{2020 IEEE/RSJ International Conference on Intelligent Robots and Systems (IROS)}, pp.\  5548--5555. IEEE, 2020.

\bibitem[Schulman et~al.(2017)Schulman, Wolski, Dhariwal, Radford, and Klimov]{schulman2017proximal}
Schulman, J., Wolski, F., Dhariwal, P., Radford, A., and Klimov, O.
\newblock Proximal policy optimization algorithms.
\newblock \emph{CoRR}, abs/1707.06347, 2017.
\newblock URL \url{http://arxiv.org/abs/1707.06347}.

\bibitem[Sutton \& Barto(2018)Sutton and Barto]{rl_is_hard}
Sutton, R.~S. and Barto, A.~G.
\newblock \emph{Reinforcement Learning: An Introduction}.
\newblock A Bradford Book, Cambridge, MA, USA, 2018.
\newblock ISBN 0262039249.

\bibitem[Tang et~al.(2023{\natexlab{a}})Tang, Lin, Akinola, Handa, Sukhatme, Ramos, Fox, and Narang]{industreal}
Tang, B., Lin, M.~A., Akinola, I., Handa, A., Sukhatme, G.~S., Ramos, F., Fox, D., and Narang, Y.
\newblock Industreal: Transferring contact-rich assembly tasks from simulation to reality, 2023{\natexlab{a}}.
\newblock URL \url{https://arxiv.org/abs/2305.17110}.

\bibitem[Tang et~al.(2023{\natexlab{b}})Tang, Lin, Akinola, Handa, Sukhatme, Ramos, Fox, and Narang]{tang2023industreal}
Tang, B., Lin, M.~A., Akinola, I., Handa, A., Sukhatme, G.~S., Ramos, F., Fox, D., and Narang, Y.
\newblock Industreal: Transferring contact-rich assembly tasks from simulation to reality.
\newblock In \emph{Robotics: Science and Systems}, 2023{\natexlab{b}}.

\bibitem[Tiboni et~al.(2024)Tiboni, Klink, Peters, Tommasi, D'Eramo, and Chalvatzaki]{entmax}
Tiboni, G., Klink, P., Peters, J., Tommasi, T., D'Eramo, C., and Chalvatzaki, G.
\newblock Domain randomization via entropy maximization, 2024.
\newblock URL \url{https://arxiv.org/abs/2311.01885}.

\bibitem[Valassakis et~al.(2020)Valassakis, Ding, and Johns]{valassakis2020crossing}
Valassakis, E., Ding, Z., and Johns, E.
\newblock Crossing the gap: A deep dive into zero-shot sim-to-real transfer for dynamics.
\newblock In \emph{2020 IEEE/RSJ International Conference on Intelligent Robots and Systems (IROS)}, pp.\  5372--5379. IEEE, 2020.

\bibitem[Wang et~al.(2025)Wang, Lv, Mao, Qu, Xu, and Ji]{wang2025robustfastadaptationadversarially}
Wang, C., Lv, Y., Mao, Y., Qu, Y., Xu, Y., and Ji, X.
\newblock Robust fast adaptation from adversarially explicit task distribution generation, 2025.
\newblock URL \url{https://arxiv.org/abs/2407.19523}.

\bibitem[Wen et~al.(2024)Wen, Yang, Kautz, and Birchfield]{wen2024foundationposeunified6dpose}
Wen, B., Yang, W., Kautz, J., and Birchfield, S.
\newblock Foundationpose: Unified 6d pose estimation and tracking of novel objects, 2024.
\newblock URL \url{https://arxiv.org/abs/2312.08344}.

\bibitem[Yu et~al.(2018)Yu, Liu, and Turk]{strategy_optimization}
Yu, W., Liu, C.~K., and Turk, G.
\newblock Policy transfer with strategy optimization.
\newblock \emph{CoRR}, abs/1810.05751, 2018.
\newblock URL \url{http://arxiv.org/abs/1810.05751}.

\bibitem[Zhang et~al.(2020)Zhang, Sharma, Liang, and Kroemer]{zhang2020modular}
Zhang, K., Sharma, M., Liang, J., and Kroemer, O.
\newblock A modular robotic arm control stack for research: Franka-interface and frankapy.
\newblock \emph{arXiv preprint arXiv:2011.02398}, 2020.

\bibitem[Zhang et~al.(2024)Zhang, Tomizuka, and Li]{dynamic_compliance}
Zhang, X., Tomizuka, M., and Li, H.
\newblock Bridging the sim-to-real gap with dynamic compliance tuning for industrial insertion, 2024.
\newblock URL \url{https://arxiv.org/abs/2311.07499}.

\bibitem[Zhu et~al.(2020)Zhu, Yu, Gupta, Shah, Hartikainen, Singh, Kumar, and Levine]{zhu2020ingredients}
Zhu, H., Yu, J., Gupta, A., Shah, D., Hartikainen, K., Singh, A., Kumar, V., and Levine, S.
\newblock The ingredients of real-world robotic reinforcement learning.
\newblock \emph{arXiv preprint arXiv:2004.12570}, 2020.

\end{thebibliography}
\bibliographystyle{icml2025}

\appendix

\newpage
\section{Appendix}

\subsection{Code Release}

The codebase for the project can be found \href{https://github.com/aidan-curtis/goflow}{here.}

\subsection{Importance Sampling Proofs}
\label{importance_sampling_proofs}

Here we show how we obtain the importance-sampled estimates for both the reward term 
\(
\mathcal{R}
\)
and the entropy term 
\(
\hat{\mathcal{H}}
\)
in \OursNoSpace (Algorithm~\ref{alg:goflow}).  
Specifically, we sample from a uniform distribution \(u(\xi)\) over \(\Xi\), rather than from the learned distribution \(p_\phi(\xi)\) directly, in order to avoid collapse onto narrow regions of the parameter space.

\subsubsection{Reward Term: \(\mathcal{R}\)}
We want an unbiased estimate of
\[
\mathbb{E}_{\xi \sim p_\phi(\xi)}\bigl[J_\xi(\pi)\bigr]
\,=\;
\int_{\Xi} 
p_\phi(\xi)\,J_\xi(\pi)
\;\mathrm{d}\xi.
\]
Let \(u(\xi)\) be a uniform distribution over \(\Xi\). Since
\(
p_\phi(\xi)\,J_\xi(\pi)
=
\tfrac{p_\phi(\xi)}{u(\xi)}\,u(\xi)\,J_\xi(\pi),
\)
we can rewrite:
\[
\int_{\Xi} 
p_\phi(\xi)\,J_\xi(\pi)
\,\mathrm{d}\xi
\;=\;
\int_{\Xi}
u(\xi)\;\frac{p_\phi(\xi)}{u(\xi)}\;J_\xi(\pi)\;\mathrm{d}\xi.
\]
Hence, sampling \(\xi_i \sim u(\xi)\) and averaging \(\tfrac{p_\phi(\xi_i)}{u(\xi_i)} J_{\xi_i}(\pi)\) gives an unbiased Monte Carlo estimate of 
\(\mathbb{E}_{p_\phi}[\,J_\xi(\pi)\bigr]\).  

If \(\Xi\) has finite measure \(\lvert \Xi\rvert\), then \(u(\xi) = 1/\lvert\Xi\rvert\). Thus
\[
\frac{p_\phi(\xi)}{u(\xi)}
\;=\;
p_\phi(\xi)\,\lvert\Xi\rvert.
\]
Therefore, the empirical estimate becomes
\[
\mathcal{R}
\;=\;
\frac{1}{\,B\,}
\sum_{i=1}^B
\bigl[
\bigl(p_\phi(\xi_i)\,\lvert\Xi\rvert\bigr)\,J_{\xi_i}(\pi)
\bigr]
\]
\[
\;=\;
\frac{\lvert \Xi\rvert}{\,B\,}
\sum_{i=1}^B
p_\phi(\xi_i)\,J_{\xi_i}(\pi).
\]
where \(\{\xi_i\}_{i=1}^B \sim u(\xi)\).  This matches Line~\ref{line:reward} in Algorithm~\ref{alg:goflow}.

\subsubsection{Entropy Term: \(\hat{\mathcal{H}}\)}
Similarly, to compute the differential entropy of \(p_\phi(\xi)\), we have
\[
\mathcal{H}(p_\phi)
\;=\;
-\int_{\Xi} p_\phi(\xi)\,\log p_\phi(\xi)\,\mathrm{d}\xi.
\]
Again, we apply the same importance sampling trick via \(u(\xi)\). We write:
\[
p_\phi(\xi)\,\log p_\phi(\xi)
\;=\;
\frac{p_\phi(\xi)}{\,u(\xi)\!}\,u(\xi)\,\log p_\phi(\xi).
\]
Hence
\[
\int_{\Xi} p_\phi(\xi)\,\log p_\phi(\xi)\,\mathrm{d}\xi
\;=\;
\int_{\Xi} u(\xi)\,\tfrac{p_\phi(\xi)}{u(\xi)}\,\log p_\phi(\xi)\,\mathrm{d}\xi.
\]
If \(\lvert \Xi\rvert\) is the measure of \(\Xi\) under \(u(\xi)\), then \(u(\xi)=1/\lvert\Xi\rvert\). So the Monte Carlo estimate for 
\(\int p_\phi(\xi)\,\log p_\phi(\xi)\,\mathrm{d}\xi\)
becomes
\[
\frac{\lvert \Xi\rvert}{\,B\,}\,
\sum_{i=1}^B
p_\phi(\xi_i)\,\log p_\phi(\xi_i),
\]
with \(\xi_i \sim u(\xi)\).  
Multiplying by \(-1\) yields the differential entropy:
\[
\mathcal{H}(p_\phi)
\;=\;
-\int p_\phi(\xi)\,\log p_\phi(\xi)\,\mathrm{d}\xi
\]

\[
\;\;\approx\;\;
-\,\frac{\lvert \Xi\rvert}{B}\,
\sum_{i=1}^B
p_\phi(\xi_i)\,\log p_\phi(\xi_i),
\]
which is exactly what we implement in Line~\ref{line:entropy} of Algorithm~\ref{alg:goflow}.

\begin{remark}
Using uniform sampling \(u(\xi)\) to approximate these terms provides global coverage of \(\Xi\), helping prevent the learned distribution \(p_\phi\) from collapsing around a small subset of parameter space.  By contrast, if one sampled \(\xi\) from \(p_\phi(\xi)\) itself for these terms, the distribution might fail to expand to other promising regions once it becomes peaked.
\end{remark}

\subsection{Domain Randomization Parameters}
\label{app:dr_params}
Below we describe the randomization ranges and parameter names for each environment. 
We also provide the reward success threshold ($J_T$) and cut the max duration of some environments in order to speed up training ($t_{max}$). 
$J_t$ was chosen to be below the optimal performance under no environment randomization. We verified that the trained policy still exhibited qualitatively successful performance at the target reward threshold. 
Lastly, we slightly modified the Quadruped environment to only take a fixed forward command rather than the goal-conditioned policy learned by default. 
Other than those changes, the first four simulated environments official IsaacLab implementation.  
\begin{itemize}
    \item \textbf{Cartpole parameters ($J_T=50, t_{max}=2s$):}
    \begin{itemize}
        \item Pole mass: Min = 0.01, Max = 20.0
        \item Cart mass: Min = 0.01, Max = 20.0
        \item Slider-Cart Friction: Min Bound = 0.0, Max Bound = 1.0
    \end{itemize}
    
    \item \textbf{Ant parameters ($J_T=700,t_{max}=2s$):}
    \begin{itemize}
        \item Torso mass: Min = 0.01, Max = 20.0
    \end{itemize}
    
    \item \textbf{Quadcopter parameters ($J_T=15, t_{max}=2s$):}
    \begin{itemize}
        \item Quadcopter mass: Min = 0.01, Max = 20.0
    \end{itemize}
    
    \item \textbf{Quadruped parameters ($J_T=1.5,t_{max}=5s$):}
    \begin{itemize}
        \item Body mass: Min = 0.0, Max = 200.0
        \item Left front hip friction: Min = 0.0, Max = 0.1
        \item Left back hip friction: Min = 0.0, Max = 0.1
        \item Right front hip friction: Min = 0.0, Max = 0.1
        \item Right back hip friction: Min = 0.0, Max = 0.1
    \end{itemize}

    \item \textbf{Humanoid parameters ($J_T=1000, t_{max}=5s$):}
    \begin{itemize}
        \item Torso Mass: Min = 0.01, Max = 25.0
        \item Head Mass: Min = 0.01, Max = 25.0
        \item Left Hand Mass: Min = 0.01, Max = 30.0
        \item Right Hand Mass: Min = 0.01, Max = 30.0
    \end{itemize}
    
    \item \textbf{Gear parameters ($J_T=50, t_{max}=4s$):}
    \begin{itemize}
        \item Grasp Pose x: Min = -0.05, Max = 0.05
        \item Grasp Pose y: Min = -0.05, Max = 0.05
        \item Grasp Pose yaw: Min = -0.393, Max = 0.393
    \end{itemize}
\end{itemize}

\subsection{Multi-Step Planning Details}
\label{app:multi-step-planning}

\subsubsection{Updating beliefs via probabilistic pose estimation}
\label{app:belief_update}

Updating the belief state $b$ requires a probabilistic state estimation system that outputs a posterior over the state space $S$, rather than a single point estimate. 
Given a new observation $o$, we use a probabilistic object pose estimation framework (Bayes3D) to infer posterior distributions over object pose~\citep{gothoskar2023bayes3d}.

The pose estimation system uses inference in an probabilistic generative graphics model with uniform priors on the translational $x$, $y$, and rotational yaw (or $r_x$) components of the 6-dof pose (since the object is assumed to be in flush contact with the table surface) and an image likelihood $P(o_\text{rgbd}\mid r_x, x, y)$. The object's geometry and color information is given by a mesh model. 
The image likelihood is computed by rendering a latent image $im^{\text{rgbd}}$ with the object pose corresponding to $(r_x,x,y)$ and calculating the per-pixel likelihood:
\begin{equation}
\begin{split}
& P(o_\text{rgbd} \mid r_x, x, y) \propto \\ 
& \prod_{i, j \in C}  \left[ p_{\text{out}} + (1-p_{\text{out}}) \cdot  P_{\text{in}}(o^\text{rgbd}_{i,j}\mid r_x, x, y) \right]
\end{split}
\end{equation}
\begin{equation}
\begin{split}
& P_{\text{in}}(o^\text{rgbd}_{i,j}\mid r_x, x, y) \propto \\
& \exp \Bigg( 
-\frac{|| o^\text{rgb}_{i,j}-im^\text{rgb}_{i,j} ||_1}{b^\text{rgb}}
- \frac{||o^\text{d}_{i,j}-im^\text{d}_{i,j}||_1}{b^\text{d}} \Bigg)
\end{split}
\end{equation}

 where $i$ and $j$ are pixel row and column indices, $C$ is the set of valid pixels returned by the renderer, $b_\text{rgb}$ and $b_\text{d}$ are hyperparameters that control posterior sensitivity to the color and depth channels, and $p_{\text{out}}$ is the pixel outlier probability hyperparameter. For an observation $o_\text{rgbd}$, we can sample from $ P(r_x, x, y\mid o_\text{rgbd}) \propto P(o_\text{rgbd}\mid r_x, x, y)$ to recover the object pose posterior with a tempering exponential factor $\alpha$ to encourage smoothness. 
 We first find the maximum a posteriori (MAP) estimate of object pose using coarse-to-fine sequential Monte Carlo sampling ~\citep{del2006sequential} and then calculate a posterior approximation using a grid centered at the MAP estimate. 
 
 The benefit of this approach in contrast to traditional rendering-based pose estimation systems, such as those presented in  \cite{wen2024foundationposeunified6dpose} or \cite{megapose}, is that our pose estimates indicate high uncertainty for distant, small, occluded, or non-visible objects as well dimensions along which the object is symmetric. 
 A visualization of the pose beliefs at different points in the multi-step plan can be seen in Figure~\ref{fig:belief_posteriors} in the Appendix.

\subsection{Hyperparameters}
\label{app:hyperparameters}
Below we list out the significant hyperparameters involved in each baseline method, and how we chose them based on our hyperparameter search. 
We run the same seed for each hyperparameter and pick the best performing hyperparameter as the representative for our larger quantitative experiments in figure~\ref{fig:sim_results}. 
The full domain randomization (FullDR) and no domain randomization (NoDR) baselines have no hyperparameters.

\subsubsection{GoFlow}
We search over the following values of the $\alpha$ hyperparameter: $[0.1, 0.5, 1.0, 1.5, 2.0]$. 
We search over the following values of the $\beta$ hyperparameters $[0.0, 0.1, 0.5, 1.0, 2.0]$.
Other hyperparameters include number of network updates per training epoch ($K=100$), network learning rate ($\eta_\phi=1e-3$), and neural spline flow architecture hyperparameters such as network depth ($\ell=3$), hidden features (64), and number of bins (8). We implement our flow using the Zuko normalizing flow library~\cite{rozet2022zuko}.

\subsubsection{LSDR}
Similary to \Ours, we search over the following values of the $\alpha_L$ hyperparameter: $[0.1, 0.5, 1.0, 1.5, 2.0]$. 
Other hyperparameters include the number of updates per training epoch (T=100), and initial Gaussian parameters: $\mu =  (\xi_{max}+\xi_{min})/2.0$ and $\Sigma = \text{diag}\left(\xi_{\text{max}} - \xi_{\text{min}}/10\right)$

\subsubsection{DORAEMON}
We search over the following values of the $\epsilon_D$ hyperparameter: $[0.005, 0.01, 0.05, 0.1, 0.5]$. 
After fixing the best $\epsilon_D$ for each environment, we additionally search over the success thrshold $\alpha_D$: $[0.005, 0.01, 0.05, 0.1, 0.5]$.

\subsubsection{ADR}
In ADR, we fix the upper threshold to be the success threshold $t^{+} = J_T$ as was done in the original paper and search over the lower bound threshold $t^{-} = [0.1t^{+}, 0.25t^{+}, 0.5t^{+}, 0.75t^{+}, 0.9t^{+}]$.
The value used in the original paper was $0.5t_H$. 
Other hyperparameters include the expansion/contraction rate, which we interpret to be a fixed fraction of the domain interval, $\Delta = 0.1 * \big[\xi_{\text{max}} - \xi_{\text{min}}\big]$, and boundary sampling probability $p_b=0.5$.

\subsection{Coverage vs Range Experiments}
\label{app:cov_vs_range}
\begin{figure*}[h!]
    \centering
    \includegraphics[width=1.0\textwidth]{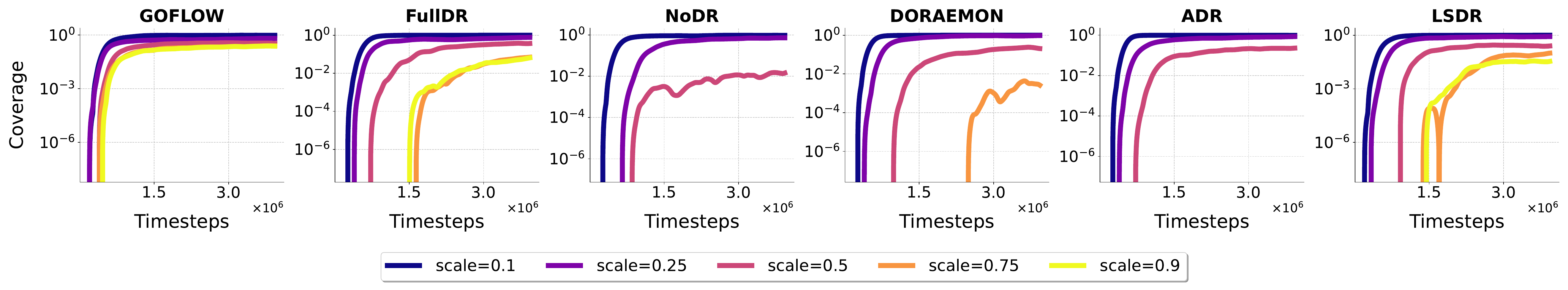}
    \caption{Coverage vs range experiment results discussed in Section~\ref{app:cov_vs_range}}
    \label{fig:scales}
\end{figure*}

We compare coverage vs. range scale in the ant domain. We adjust the parameter lower and upper bounds outlined in Appendix~\ref{app:dr_params} and see how the coverage responds to those changes during training. 
The parameter range is defined relative to a nominal midpoint $m$ set to the original domain parameters: $[m - (m - \text{lower}) * \text{scale}, m + (\text{upper} - m) * \text{scale})]$. 
The results of our experiment are shown in Figure~\ref{fig:scales}

\label{app:coverage_experiments}

\subsection{Coverage vs. Threshold Experiments}
\label{app:cov_vs_thresh}
\begin{figure*}[h!]
    \centering
    \includegraphics[width=1.0\textwidth]{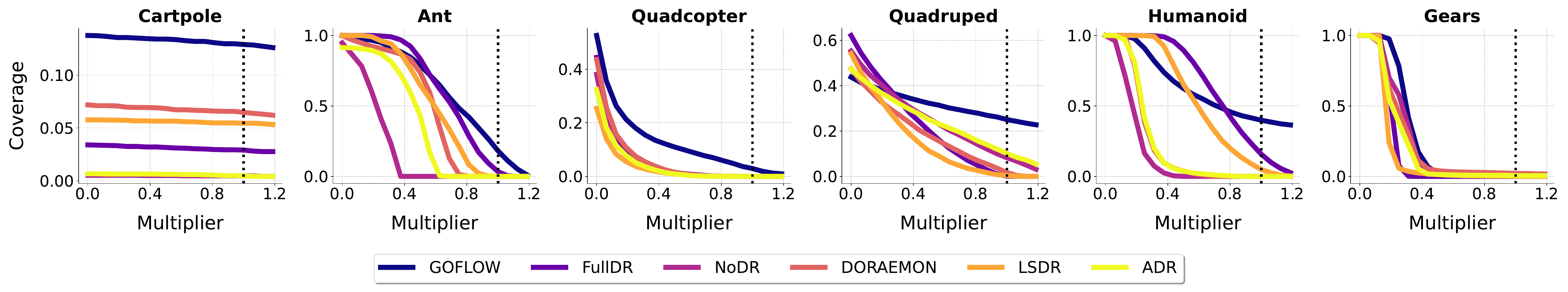}
    \caption{Coverage vs threshold experiment results discussed in Section~\ref{app:cov_vs_thresh}}
    \label{fig:thresholds}
\end{figure*}

In this experiment, we assess the sensitivity of coverage to the performance threshold $J_t$ for a single training seed. We experiment with threshold multipliers between $0$ and $1.2\cdot{J_t}$. The results of this experiment can be seen in Figure~\ref{fig:thresholds}.
These results demonstrate that although coverage is highly dependent on $J_t$, \Ours largely dominates the coverage distribution for wide ranges of $J_t$.
\label{app:threshold_experiments}

\subsection{Real-world experiments}

In addition to simulated experiments, we compare GoFlow against baselines on a real-world gear insertion task. 
In particular, we tested insertion of the pink medium gear over 10 trials for each baseline. 
To test this, we had the robot perform 10 successive pick/inserts of the pink gear into the middle shaft of the gear plate. Instead of randomizing the pose of the gear, we elected to fix the initial pose of the gear and the systematically perturb the end-effector pose by a random $\pm 0.01$ meter translational offset along the x dimension during the pick. 
We expect some additional grasp pose noise due to position error during grasp and object shift during grasp. 
This led to a randomized in-hand offset while running the trained insertion policy. 
Our results show that GoFlow can indeed more robustly generalize to real-world robot settings under pose uncertainty.

\subsection{Statistical Tests}
We performed a statistical analysis of the simulated results reported in Figure~\ref{fig:sim_results} and the real-world experiments in Table~\ref{tab:real_world}. For the simulated results, we recorded the final domain coverages across all seeds and performed pairwise t-tests between each method and the top-performing method. 
The final performance mean and standard deviation are reported in Table~\ref{tab:statistical}. Any methods that were not significantly different from the top performing method ($p<0.05$) are bolded. 
This same method was used to test significance of the real-world results.

\newpage

\begin{figure*}[h!]
    \centering
    \includegraphics[width=\linewidth]{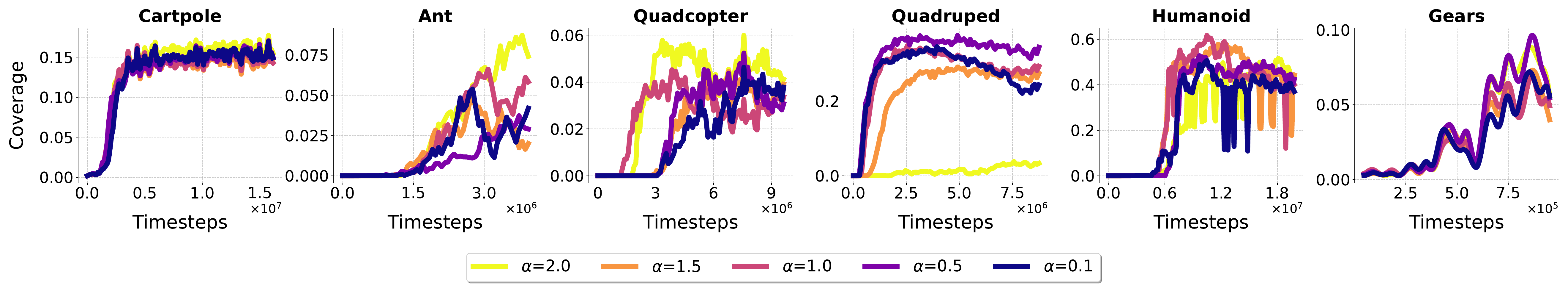}
    \caption{ \Ours hyperparameter sweep results for $\alpha$}
    \label{fig:goflow_alpha_results}
\end{figure*}

\begin{figure*}[h!]
    \centering
    \includegraphics[width=\linewidth]{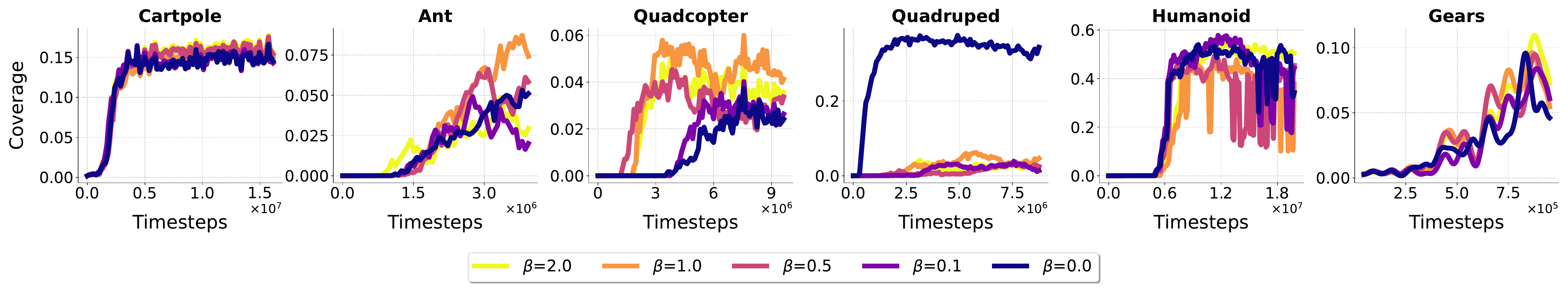}
    \caption{\Ours hyperparameter sweep results for $\beta$ after fixing the best $\alpha$ for each environment}
    \label{fig:goflow_beta_results}
\end{figure*}

\begin{figure*}[h!]
    \centering
    \includegraphics[width=\linewidth]{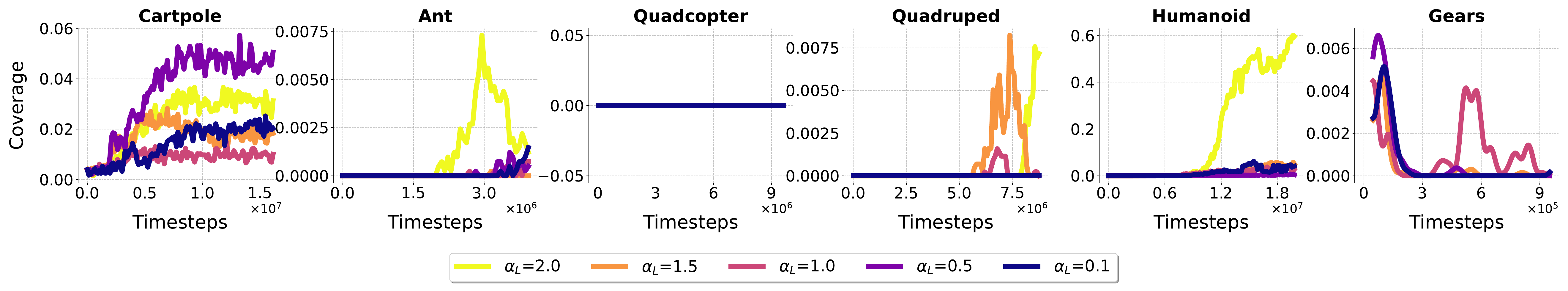}
    \caption{LSDR hyperparameter sweep results for $\alpha_L$}
    \label{fig:lsdr_alpha_results}
\end{figure*}

\begin{figure*}[h!]
    \centering
    \includegraphics[width=\linewidth]{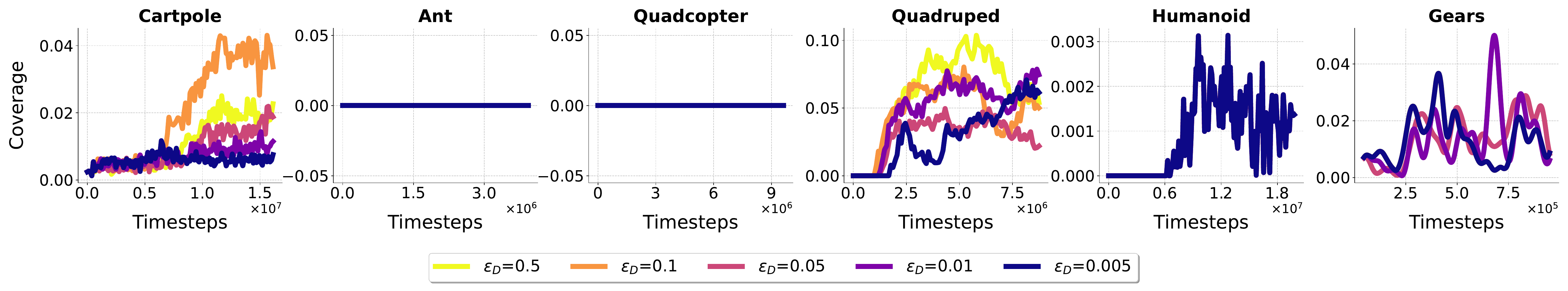}
    \caption{DORAEMON hyperparameter sweep results for $\epsilon_D$}
    \label{fig:doraemon_epsilon_results}
\end{figure*}

\begin{figure*}[h!]
    \centering
    \includegraphics[width=\linewidth]{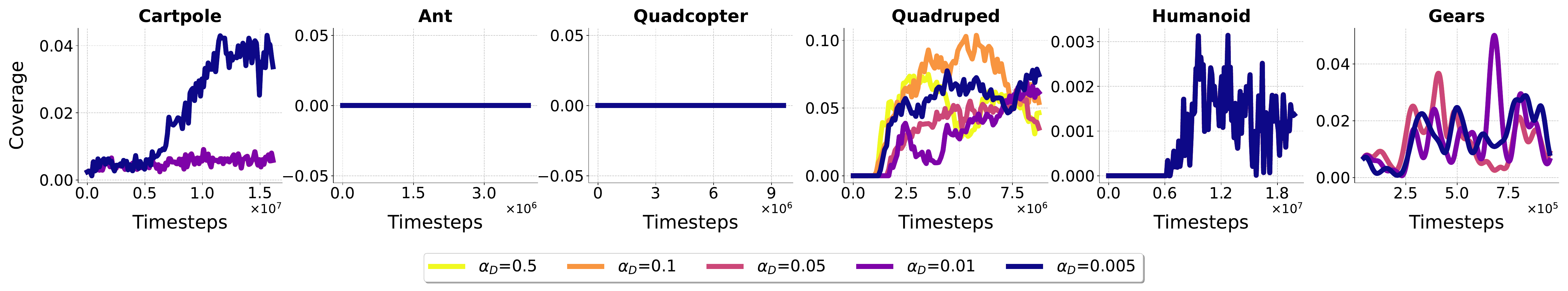}
    \caption{DORAEMON hyperparameter sweep results for $\alpha_D$ after fixing the best $\epsilon_D$}
    \label{fig:doraemon_alpha_results}
\end{figure*}

\begin{figure*}[h!]
    \centering
    \includegraphics[width=\linewidth]{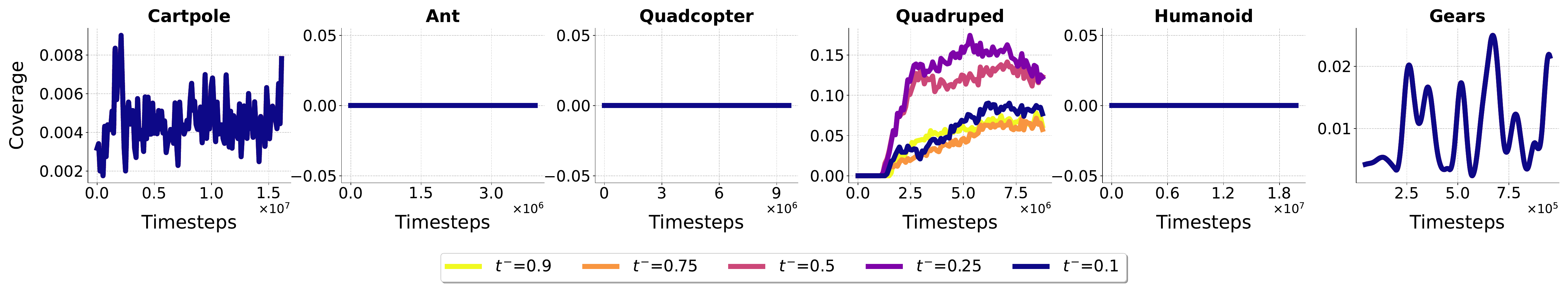}
    \caption{ADR hyperparameter sweep results for $t^{-}$}
    \label{fig:adr_t_results}
\end{figure*}

\begin{table*}[h!]

    \centering
    \setlength{\tabcolsep}{8pt} 
    \renewcommand{\arraystretch}{1.5} 
    \begin{tabular}{|l|l|l|l|l|l|l|}
        \hline
        Method   & FullDR & NoDR & DORAEMON & LSDR & ADR & GoFlow \\ \hline
        Success Rate & 6/10 & 3/10 & 5/10 & 5/10 & 5/10 & \textbf{9/10} \\ \hline
    \end{tabular}
    \caption{Real-world experimental results with the statistically significant results bolded.}
    \label{tab:real_world}

\end{table*}

\begin{table*}[h]
    \centering
    \setlength{\tabcolsep}{5pt} 
    \begin{tabular}{|l|c|c|c|c|c|c|}
        \hline
         & Cartpole & Ant & Quadcopter & Quadruped & Humanoid & Gears \\ \hline
        FullDR & 0.060$\pm$0.021 & \textbf{0.014$\pm$0.006} & 0.000$\pm$0.000 & 0.012$\pm$0.007 & \textbf{0.195$\pm$0.044} & 0.000$\pm$0.000 \\ \hline
        NoDR & 0.005$\pm$0.001 & 0.000$\pm$0.000 & 0.000$\pm$0.000 & 0.086$\pm$0.008 & 0.001$\pm$0.001 & \textbf{0.008$\pm$0.003} \\ \hline
        DORAEMON & 0.015$\pm$0.005 & 0.000$\pm$0.000 & 0.000$\pm$0.000 & 0.029$\pm$0.008 & 0.000$\pm$0.000 & \textbf{0.012$\pm$0.004} \\ \hline
        LSDR & 0.103$\pm$0.013 & 0.000$\pm$0.000 & \textbf{0.007$\pm$0.003} & 0.009$\pm$0.008 & \textbf{0.176$\pm$0.082} & 0.000$\pm$0.000 \\ \hline
        ADR & 0.006$\pm$0.001 & 0.000$\pm$0.000 & 0.000$\pm$0.000 & 0.070$\pm$0.006 & 0.000$\pm$0.000 & \textbf{0.008$\pm$0.004} \\ \hline
        GOFLOW & \textbf{0.138$\pm$0.005} & \textbf{0.056$\pm$0.033} & \textbf{0.010$\pm$0.006} & \textbf{0.275$\pm$0.019} & \textbf{0.274$\pm$0.076} & \textbf{0.035$\pm$0.019} \\ \hline
    \end{tabular}
    \caption{Mean and standard error (SDE) of the final reward value, with entries in bold indicating the best-performing method or methods not statistically significantly worse than the best (one-tailed z test, $\alpha=0.05$).}
    \label{tab:statistical}
\end{table*}

\begin{table*}[h]
    \centering
    \setlength{\tabcolsep}{5pt} 
    \begin{tabular}{|l|c|c|c|c|c|c|}
        \hline
         & Cartpole & Ant & Quadcopter & Quadruped & Humanoid & Gears \\ \hline
        FullDR & -920.132 & 362.454 & -2.286 & 0.108 & 657.414 & 8.627 \\ \hline
        NoDR & -949.925 & 133.331 & -3.807 & -4.062 & 180.734 & 12.697 \\ \hline
        DORAEMON & -930.417 & 86.742 & -1.767 & -2.472 & 235.516 & 11.378 \\ \hline
        LSDR & -867.347 & 367.612 & -1.870 & -0.052 & 532.531 & 9.289 \\ \hline
        ADR & -953.498 & 64.815 & -3.183 & -2.926 & 246.297 & 9.671 \\ \hline
        GOFLOW & -820.510 & 198.302 & -1.000 & -2.814 & 376.652 & 14.209 \\ \hline
    \end{tabular}
    \caption{CVaR computed as the mean of the final rewards falling below the 10\% percentile (VaR).}
    \label{tab:cvar}
\end{table*}

\begin{figure*}[h]
    \centering
    \includegraphics[width=1.0\textwidth]{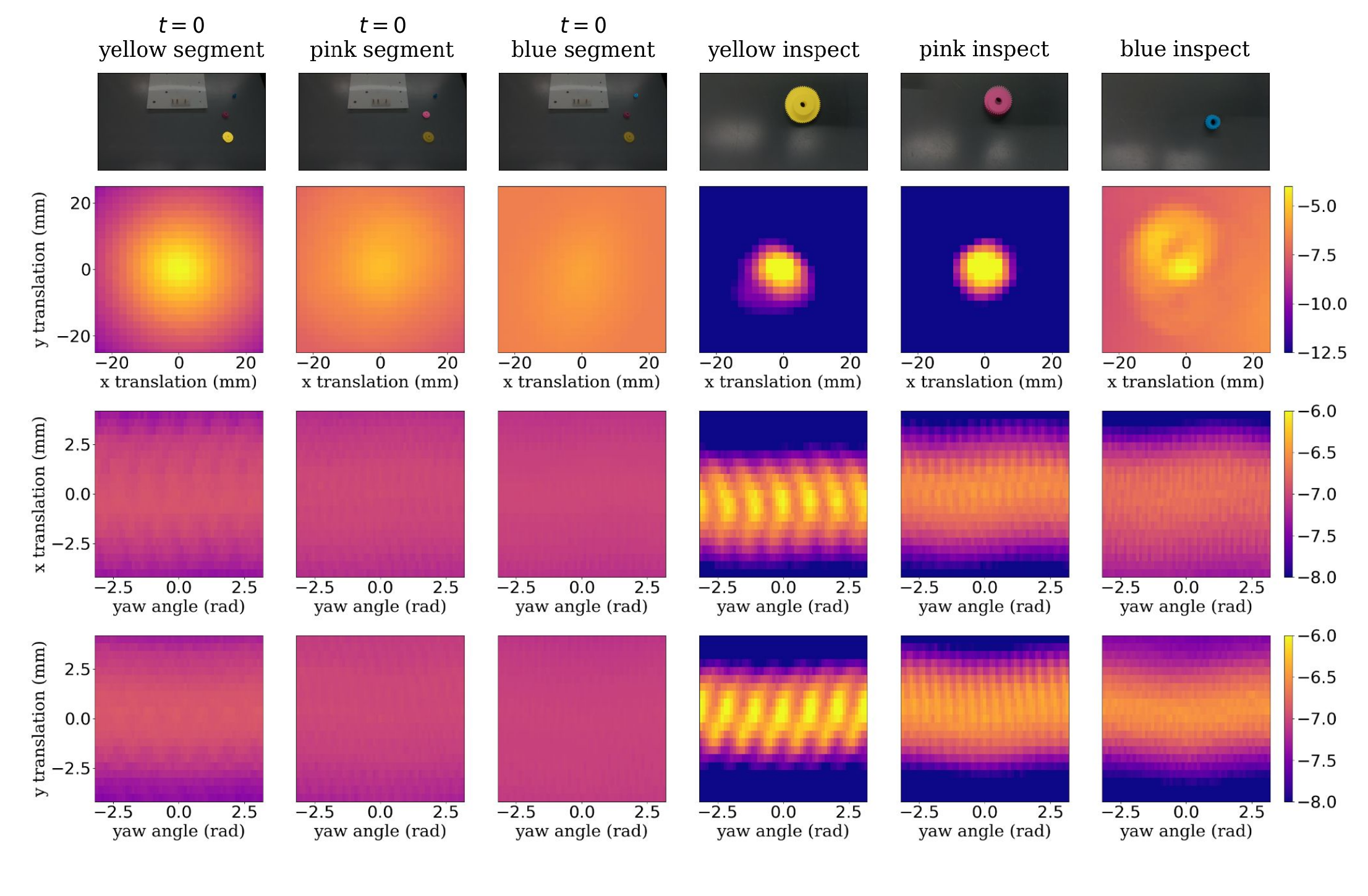}
    \caption{A visualization of the beliefs over the object pose under the initial image (first three columns) and after closer inspection (last three columns) as generated from the posterior of the model described in Section~\ref{sec:belief_update}. The colormap corresponds to the log probability of the posterior pose estimate. All plots are centered around the most likely pose estimate under the image model. }
    \label{fig:belief_posteriors}
\end{figure*}

\begin{algorithm*}
\caption{Belief-Space Planner Using BFS}
\label{alg:belief_space_planner}
\begin{algorithmic}[1]
\Require Initial belief state \( b_0 \), goal condition \( G \subseteq \mathcal{B} \), set of skills \( \mathcal{A}_\Pi \), success threshold \( \eta \)
\State Initialize the frontier \( \mathcal{F} \gets \{ b_0 \} \)
\State Initialize the visited set \( \mathcal{V} \gets \emptyset \)
\State Initialize the plan dictionary \( \text{Plan} \) mapping belief states to sequences of skills
\While{ \( \mathcal{F} \) is not empty }
    \State Dequeue \( b \) from \( \mathcal{F} \)
    \If{ \( b \in G \) }
        \State \Return \( \text{Plan}[b] \) \Comment{Return the sequence of skills leading to \( b \)}
    \EndIf
    \ForAll{ skills \( \pi \in \mathcal{A}_\Pi \) }
        \If{ \( b \in \text{Pre}_\pi \) given \( \eta \) }
            \State \( b' \gets \texttt{sample}(\text{Eff}_\pi) \) 
            \If{ \( b' \notin \mathcal{V} \) }
                \State Add \( b' \) to \( \mathcal{F} \) and \( \mathcal{V} \)
                \State Update \( \text{Plan}[b'] \gets \text{Plan}[b] + [\pi] \)
            \EndIf
        \EndIf
    \EndFor
\EndWhile
\State \Return \textbf{Failure} \Comment{No plan found}
\end{algorithmic}
\end{algorithm*}

\end{document}